\documentclass[letterpaper]{article}
\usepackage[preprint]{aaai2027}
\usepackage[hyphens]{url}  \usepackage{graphicx}    \usepackage{natbib}  \usepackage{caption} 
\usepackage{algorithm}
\usepackage{algorithmic}
\usepackage{array}
\usepackage{booktabs}
\usepackage{colortbl}
\usepackage{tabularx}
\usepackage{multirow}
\usepackage{arydshln}
\usepackage{cuted}
\usepackage{pifont}
\usepackage{needspace}
\usepackage[most]{tcolorbox}

\definecolor{bdblue}{HTML}{2F6BFF}
\definecolor{bdcyan}{HTML}{00C2FF}
\definecolor{bdorange}{HTML}{E69F00}
\definecolor{bdred}{HTML}{FE2C55}
\definecolor{bdink}{HTML}{111827}
\definecolor{bdline}{HTML}{BFD4FF}
\definecolor{bdheader}{HTML}{EAF3FF}
\definecolor{bdstripe}{HTML}{F6FBFF}
\definecolor{bdwin}{HTML}{E8F7FF}
\definecolor{bdsoftred}{HTML}{FFF0F3}
\definecolor{bdteal}{HTML}{007F7A}
\definecolor{bdtealLine}{HTML}{82C7C2}
\definecolor{bdtealHeader}{HTML}{EAF7F6}
\definecolor{bdtealStripe}{HTML}{F7FBFB}
\definecolor{bdtealWin}{HTML}{E7F8FA}
\definecolor{bdexroleHeader}{HTML}{B8E8D2}
\definecolor{bdexroleRow}{HTML}{D9F4E7}
\definecolor{bdexroleGray}{HTML}{F1F3F5}
\definecolor{bdgreen}{HTML}{00A76F}
\definecolor{bdgold}{HTML}{FFE8A3}
\definecolor{bdneutral}{HTML}{667085}
\newcolumntype{L}[1]{>{\raggedright\arraybackslash}p{#1}}
\newcolumntype{C}[1]{>{\centering\arraybackslash}p{#1}}

\newcommand{\negdelta}[1]{\textcolor{bdred}{\tiny (#1)}}

\newcommand{\metricval}[2]{\mbox{#1\raisebox{-0.35ex}{\,#2}}}

\newcommand{\mneg}[2]{\metricval{#1}{\negdelta{#2}}}

\newcommand{\bestplain}[1]{\underline{\textbf{\textit{#1}}}}

\newcommand{\metricup}{\ensuremath{\uparrow}}
\newcommand{\basesft}{\textcolor{bdteal}{\ding{72}}}
\newcommand{\baseqwen}{\textcolor{bdblue}{\ding{117}}}
\newcommand{\baseragrl}{\textcolor{bdred}{\ding{108}}}
\newcommand{\alginputicon}{\textcolor{bdblue}{\ding{43}}}
\newcommand{\algoutputicon}{\textcolor{bdgreen}{\ding{51}}}
\newcommand{\algstageicon}{\textcolor{bdteal}{\ding{117}}}

\newcommand{\titleicon}{\textcolor{bdblue}{\ding{72}}}
\newcommand{\appanchor}[1]{\pdfdest name {#1} xyz\relax}
\newcommand{\appjump}[2]{\leavevmode\pdfstartlink attr {/Border [0 0 0]} goto name {#1}\relax #2\pdfendlink}
\newcommand{\authorlogo}[2]{\raisebox{-0.18em}{\includegraphics[height=#1]{#2}}}

\title{\titleicon\ ExRole: From Team Trajectories to Executable Roles\\
in Multi-Agent Language Models}

\author{
  Zhou Liu\textsuperscript{\rm 1},
  Chaoyang Han\textsuperscript{\rm 1},
  Zewei Pan\textsuperscript{\rm 2},
  Zeli Su\textsuperscript{\rm 1},
  Wentao Zhang\textsuperscript{\rm 1}\corresponding
}

\affiliations{
  \textsuperscript{\rm 1}\authorlogo{1.05em}{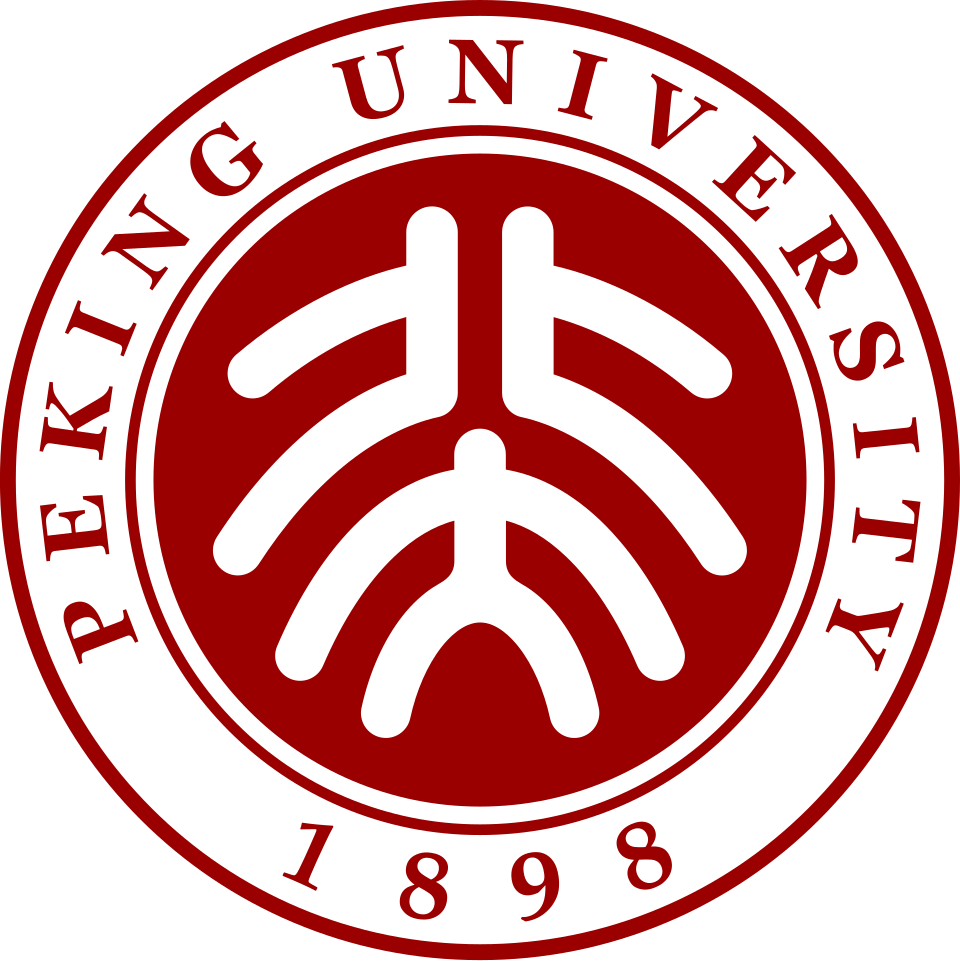}\ Peking University
  \quad
  \textsuperscript{\rm 2}\authorlogo{1.05em}{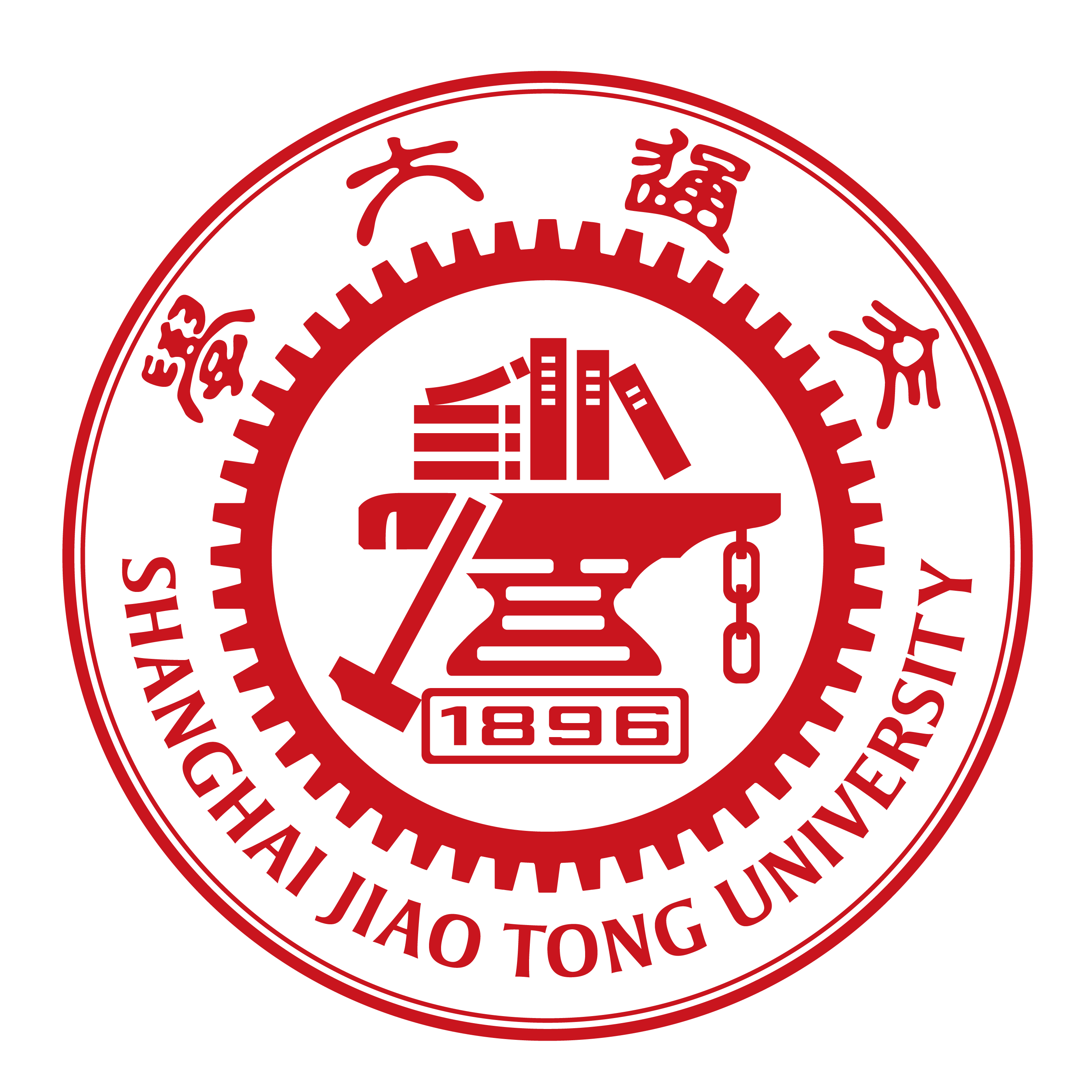}\ Shanghai Jiao Tong University
}

\begin{document}

\maketitle

\begin{abstract}
Roles provide an interpretable interface for organizing language-model agents, yet most multi-agent systems treat them as hand-written prompt labels disconnected from learned behavior and parameter updates.  We argue that a useful role should instead be an executable control variable: it should summarize behavior predictive of future utility, guide subsequent interaction, and identify the trainable capacity responsible for that behavior.  We introduce ExRole, a trajectory-to-role framework that learns future-aware role prototypes from prefix-local team traces, resolves them into readable instructions and token-aligned role markers, and optionally routes shared LoRA rank slots with turn-aligned credit.  Across MuSiQue and 2WikiMultiHopQA, ExRole improves over single-agent search by 15.0/14.4 and 13.5/16.1 EM/F1 points, respectively.  Against the strongest non-ExRole controls, the corresponding gains remain 11.5/11.6 and 7.7/9.7 points.  Across both benchmarks, the controlled results consistently favor trajectory-induced role conditioning over role-free, manual, random, and shuffled alternatives.  Role-Agent-Turn interventions further show that the induced roles capture transferable behavioral specialization beyond fixed agent identities or turn positions.
\end{abstract}

\section{Introduction}

\begin{figure}[!t]
\centering
\includegraphics[width=0.9\linewidth]{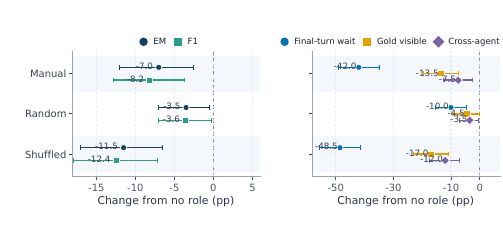}
\begin{minipage}[t]{0.48\linewidth}
\centering\small\textbf{(a)} Answer quality
\end{minipage}\hfill
\begin{minipage}[t]{0.48\linewidth}
\centering\small\textbf{(b)} Evidence-use behavior
\end{minipage}
\caption{Role-prompt interventions under a fixed MuSiQue policy. Points show paired changes from no role over the same 200 questions; bars denote 95\% paired-bootstrap confidence intervals.}
\label{fig:role-prompt-intervention}
\end{figure}

Large language model (LLM) agents increasingly solve knowledge-intensive tasks by interleaving reasoning with retrieval~\cite{trivedi2023ircot,jin2025searchr1}; recent multi-agent search work further distributes planning, retrieval, and synthesis across specialized agents~\cite{chen2026mhgpo}.  Roles offer a simple and interpretable abstraction for organizing this collaboration: they expose the intended division of labor without requiring a separate model for every agent.  Role-playing designs commonly instantiate these functions as natural-language profiles fixed in prompts before execution, as in {CAMEL}'s inception prompting~\cite{li2023camel}.  Figure~\ref{fig:role-prompt-intervention} illustrates that changing such labels alone does not reliably improve answer quality or evidence use under a fixed policy.  These labels are neither grounded in observed team behavior nor tied to the learning signal for role turns.

\noindent We ask: how can a system induce roles from prior collaboration and reuse them to coordinate new episodes and assign policy credit?

Learning roles from collaboration trajectories is more demanding than clustering surface-level action counts.  First, a role representation must be inferred from information available at the current prefix while still capturing how that behavior affects future evidence and return; otherwise it either ignores delayed utility or leaks suffix information.  Second, an induced cluster must become executable in a new episode: the same identity should guide the agent's natural-language instruction and remain recoverable inside the model.  Third, shared team rewards create the familiar credit-assignment problem of identifying which action produced the outcome~\cite{foerster2018coma}; here the credit must further reach the role turn and capacity path that produced a useful textual action.  Recent role-decomposed and segment-typed agentic RL methods likewise show why uniform outcome credit can obscure useful intermediate contributions~\cite{park2026dac,xu2026triage}.  Without these links, role induction, role prompting, and policy optimization remain separate mechanisms rather than a learned specialization process.

Our key insight is to treat a role as a shared control variable that connects behavioral abstraction with policy execution.  We instantiate this idea in ExRole, an executable role-learning framework illustrated in Figure~\ref{fig:exrole-overall-pipeline}.  ExRole first encodes prefix-local action, timing, interaction, and evidence statistics, and uses next-action, future-evidence, and final-return targets to organize trajectories by their prospective utility.  It then summarizes each cluster as a reusable prototype and deterministically resolves the prototype into a readable instruction and a token-aligned role marker, without requiring an additional role-labeling language model.

The resolved identity controls two complementary paths.  At the interaction level, the instruction conditions the active agent and the marker remains aligned with the generated role-turn tokens.  At the parameter level, ExRole-Routed combines that marker with the current semantic prefix to select a balanced sparse-delta gate over shared LoRA rank slots.  GRPO is augmented with turn-aligned credit so that a useful role turn reinforces both its response tokens and, when routing is enabled, its selected capacity.  ExRole-Shared retains the same induced roles and turn-aligned policy credit but uses a uniform shared LoRA path, providing a matched control that isolates the contribution of role-conditioned sparse LoRA routing.

We evaluate ExRole on MuSiQue and 2WikiMultiHopQA, using single-agent search, no-role and manual-role teams, random role prompts, and shuffled induced roles to separate trajectory-derived specialization from generic prompting effects.  On MuSiQue, ExRole-Shared reaches 31.5 EM and 43.2 F1, improving single-agent search by 15.0 and 14.4 points and the strongest non-ExRole control by 11.5 and 11.6 points.  On 2WikiMultiHopQA, ExRole-Routed reaches 50.0 EM and 59.7 F1, gains of 13.5 and 16.1 points over single-agent search and 7.7 and 9.7 points over the strongest control.  Across both benchmarks, the controlled comparisons favor trajectory-induced roles over role-free, manual, random, and shuffled alternatives.  Role-Agent-Turn interventions further show that the induced roles encode transferable action tendencies beyond fixed agent identities and turn positions.

To sum up, our contributions are two-fold:
\begin{itemize}
\item We formulate LLM-agent role learning as trajectory-conditioned role induction, in which prefix-local behavior and future-utility targets define reusable role prototypes.
\item We make induced roles executable by binding each prototype to a readable instruction, a token-aligned role marker, and a deterministic agent assignment, and connect that identity to role-conditioned sparse LoRA routing with turn-aligned credit.
\end{itemize}

\section{Related Work}

\subsection{Role Discovery in Multi-Agent Reinforcement Learning}

Role-based multi-agent reinforcement learning (MARL) learns reusable abstractions through identifiable embeddings (ROMA), role-specific action spaces (RODE), trajectory-encoded abilities (LDSA), contrastive representations (ACORM), and future behavioral effects (R3DM), while retaining parameter sharing~\cite{wang2020roma,wang2021rode,yang2022ldsa,hu2024acorm,goel2025r3dm}.  These methods primarily assume compact states and actions; ExRole extends role induction to language-agent trajectories, where a role must organize messages, retrieved evidence, delayed answers, and trainable capacity.

\subsection{Roles and Credit Assignment in LLM Multi-Agent Systems}

LLM teams commonly begin with prompt-defined roles, as in CAMEL, while MLC, MasRouter, and ReSo learn role differentiation, allocation, or agent selection~\cite{li2023camel,li2025collaborative,yue2025masrouter,zhou2025reso}.  MARFT, MAGRPO, MHGPO, and MATPO instead optimize team policies within a specified organization~\cite{liao2025marft,liu2026magrpo,chen2026mhgpo,mo2025matpo}.  Retrieval-oriented reasoning adds delayed credit: IRCoT interleaves retrieval and reasoning, whereas Search-R1 and R1-Searcher optimize single search policies~\cite{trivedi2023ircot,jin2025searchr1,song2025r1searcher}.  COMA addresses team credit in MARL, and DAC and TRIAGE provide cross-agent or segment-level learning signals~\cite{foerster2018coma,park2026dac,xu2026triage}.  ExRole differs by inducing recurring functions from prior trajectories and reusing the same identity across instructions, role-turn credit, and sparse LoRA routing.

\subsection{Parameter-Efficient Specialization and Routing}

Shared-backbone systems preserve specialization through selective parameter use: Kaleidoscope learns agent-specific masks and ADMN routes agents through shared modules~\cite{li2024kaleidoscope,yu2024admn}.  LoraHub, LoraRetriever, X-LoRA, and MoLE compose or route LoRA adapters from task or input signals~\cite{huang2024lorahub,zhao2024loraretriever,buehler2024xlora,wu2024mole}.  ExRole instead derives routing identity from collaboration trajectories and modulates rank slots inside a jointly trained shared LoRA parameterization.

\section{Problem Formulation}

\begin{figure*}[t]
\centering
\includegraphics[width=0.9\textwidth]{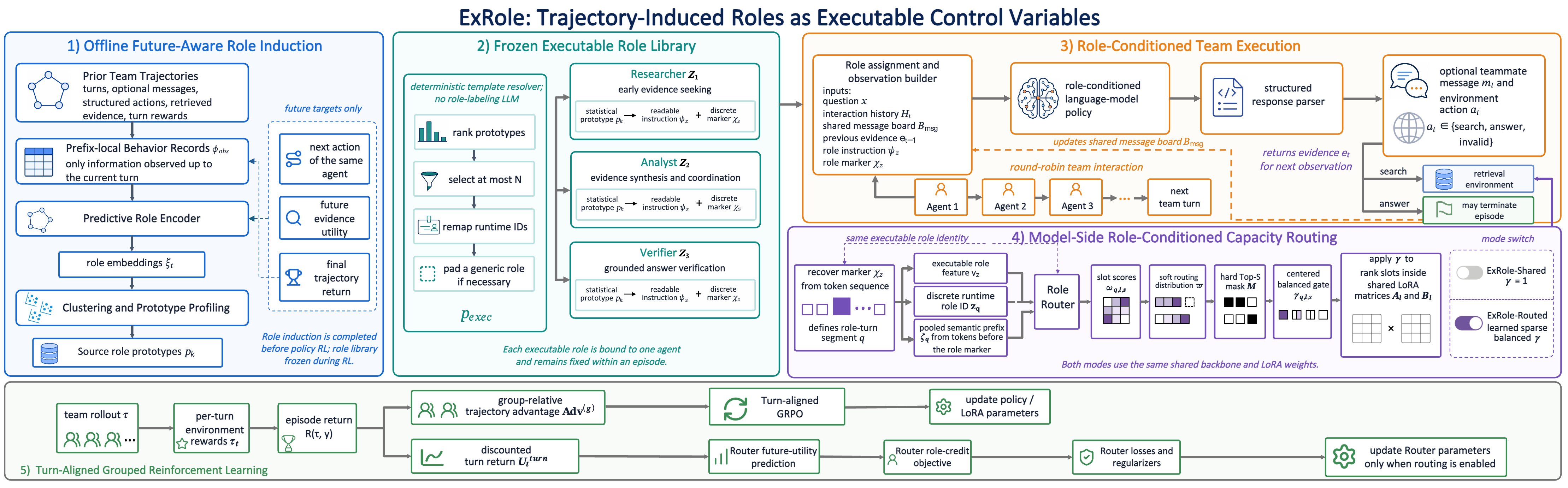}
\caption{Overview of ExRole. Prior team trajectories induce a future-aware role library. The same role identity conditions prompt-level coordination and role-conditioned sparse LoRA routing, while turn-aligned GRPO updates the policy and router.}
\label{fig:exrole-overall-pipeline}
\end{figure*}

We study role learning for multi-agent language-model teams in evidence-seeking tasks.  Each instance is a pair $(x,y)$, where $x$ is the input question and $y$ is a gold answer.  A team of $N$ agents interacts with a retrieval environment for at most $T$ turns, producing

\begin{equation}
\tau=
\left(
x,
\{i_t,o_t,m_t,a_t,e_t,r_t\}_{t=1}^{T_\tau},
\hat y
\right),
\qquad T_\tau\leq T .
\end{equation}

Here $i_t$ is the active agent, $o_t$ its observation, $m_t$ an optional teammate message, $a_t$ a search, answer, or invalid action, $e_t$ retrieved evidence, $r_t$ the turn reward, and $\hat y$ the final answer.  A team turn is one active-agent response; a role-turn segment is the token span aligned to its role marker.  Let $\mathcal H_t$ denote the interaction history available before turn $t$.  Agent $i$ is assigned an executable role $z_i$, and the active policy is conditioned on both that role and its adapter-routing state $\Gamma_t$:

\begin{equation}
(m_t,a_t)
\sim
\pi_{\theta,\theta_b}
\left(
\cdot\mid x,\mathcal H_t,z_{i_t},\Gamma_t
\right).
\end{equation}

ExRole induces a source role library $\mathcal P=\{p_k\}_{k=0}^{K-1}$ from prior trajectories and resolves it into $N$ executable roles.  The policy parameters $\theta$ and, when enabled, router parameters $\theta_b$ maximize the expected episode reward

\begin{equation}
\begin{aligned}
R(\tau,y)&=\sum_{t=1}^{T_\tau}r_t,\\
(\theta^\star,\theta_b^\star)
&=\arg\max_{\theta,\theta_b}
{\bf E}_{(x,y)}
{\bf E}_{\tau\sim
\pi_{\theta,\theta_b}(\cdot\mid x,\mathcal P_{\mathrm{exec}})}
[R(\tau,y)].
\end{aligned}
\end{equation}

The complete action grammar, reward decomposition, and answer normalizer are provided in the appendix.

\section{Method}

ExRole addresses the gap between discovering recurring team behaviors and turning them into specialization that can be executed and optimized in later episodes.  Given prior collaboration trajectories, our goal is to induce roles that capture future-relevant behavior, bind each role to a consistent interaction and model-side identity, and assign learning credit to the role and capacity path that produced an action.  Figure~\ref{fig:exrole-overall-pipeline} summarizes the resulting pipeline: predictive trajectory encoding and clustering produce a role library; deterministic role binding makes each prototype executable; role-conditioned interaction and sparse LoRA routing apply the same identity during generation; and turn-aligned grouped reinforcement learning updates the policy and, when enabled, the router.  The following sections describe these four components in order.

\subsection{Future-Aware Role Induction}

A useful role should summarize not only what an agent has done, but also what its behavior predicts about subsequent collaboration.  For each logged agent turn, ExRole constructs a prefix-local behavior vector

\begin{equation}
\begin{aligned}
\phi_t^{\mathrm{obs}}
&=
\left[
\phi_t^{\mathrm{act}},
\phi_t^{\mathrm{pos}},
\phi_t^{\mathrm{team}},
\phi_t^{\mathrm{evd}},
\phi_t^{\mathrm{ans}}
\right]\in\mathbf R^{30},\\
\widetilde\phi_{t,j}^{\mathrm{obs}}
&=
\frac{\phi_{t,j}^{\mathrm{obs}}-\mu_j}
{\sigma_j+10^{-6}}.
\end{aligned}
\end{equation}

which summarizes action tendencies, turn position, team context, evidence use, and answer behavior observed up to that point; $\mu_j$ and $\sigma_j$ are corpus statistics for coordinate $j$.  Future events are excluded from $\phi_t^{\mathrm{obs}}$ and used only as prediction targets.  A role encoder produces an embedding $\xi_t$ and predicts the same agent's next action, future evidence utility, and final trajectory return using cross-entropy (CE), binary cross-entropy (BCE), and squared-error terms:

\begin{equation}
\begin{aligned}
\xi_t&=f_{\theta_e}(\widetilde\phi_t^{\mathrm{obs}}),\\
(\hat a_t,\widehat v_t^{\mathrm{evd}},\widehat R_t)
&=h_{\theta_e}(\xi_t),\\
\mathcal L_{\mathrm{role}}
&=\mathrm{CE}(\hat a_t,a_{t_i^+})
+\lambda_{\mathrm{evd}}
\mathrm{BCE}(\widehat v_t^{\mathrm{evd}},v_t^{\mathrm{evd}})\\
&\quad
+\lambda_{\mathrm{ret}}
\left(
\widehat R_t-R(\tau(t),y_{\tau(t)})
\right)^2 .
\end{aligned}
\end{equation}

Here $t_i^+$ is the next turn of the same agent, or a stop target if no such turn exists.  This predictive objective shapes the embedding with future utility without placing suffix information in its input.

For each candidate role count, we run Euclidean K-means from each configured seed and summarize every sufficiently supported cluster as a source prototype:

\begin{equation}
\begin{aligned}
\mathbf k_K^{(s)}
&=\operatorname{KMeans}_{10}
(\{\xi_t\}_{t=1}^{J};K,s),\\
p_k&=(\bar\xi_k,\bar\phi_k,\eta_k),
\qquad k\in\mathcal C_K,\\
\mathcal C_K&=\{k:n_k\geq n_0\},
\end{aligned}
\end{equation}

where the subscript $10$ denotes ten K-means initializations, $s$ is a discovery seed, $n_k$ is cluster support, and $n_0$ is the minimum support.  The vectors $\bar\xi_k$ and $\bar\phi_k$ are cluster centroids, and $\eta_k$ records the cluster's dominant action, stage, evidence, and return statistics.  When $K$ is not fixed by the execution budget, it is selected by a separation--stability criterion,

\begin{equation}
K^\star
=
\arg\max_{K\in\mathcal K}
\left[
\overline{\mathrm{Sil}}(K)
+\lambda_{\mathrm{stab}}\operatorname{Stability}(K)
\right].
\end{equation}

The appendix defines the seed averages and adjusted-Rand stability statistic.  Finally, a deterministic joint resolver scores all surviving prototype--template pairs, assigns distinct functional types when possible, and converts the assignments into readable instructions $\psi_k$ and token-aligned role markers $\chi_k$:

\begin{equation}
\begin{aligned}
S_{k,\upsilon}
&=\mathrm{TemplateScore}_{\upsilon}(\bar\phi_k,\eta_k),\\
\{\upsilon_k\}_{k\in\mathcal C_K}
&=\mathrm{GreedyDistinct}(S;\prec_{\mathrm{tie}}),\\
(\psi_k,\chi_k)
&=\mathrm{Template}_{\upsilon_k}(\bar\phi_k,\eta_k).
\end{aligned}
\end{equation}

Thus, role names are interpretations of learned behavioral clusters rather than manually specified training labels.  Feature definitions, future-target construction, role-count settings, and template rules appear in the appendix.

\subsection{Executable Role Binding}

Induced clusters become useful only when they control subsequent team behavior.  ExRole selects at most $N$ source prototypes, remaps them to contiguous runtime role IDs, and pads missing entries with a generic collaborative role:

\begin{equation}
\mathcal P_{\mathrm{exec}}
=\mathrm{Resolve}_N(\mathcal P),
\qquad
(z_i,\psi_{z_i},\chi_{z_i})
=\mathrm{Bind}_i(\mathcal P_{\mathrm{exec}}).
\end{equation}

The resolved role conditions the observation presented to the active agent,

\begin{equation}
o_t=
\mathcal O
\left(
x,i_t,z_{i_t},\psi_{z_{i_t}},\chi_{z_{i_t}},
\mathcal H_t,\mathcal B_t^{\mathrm{msg}},e_{t-1}
\right),
\end{equation}

\noindent where $\mathcal B_t^{\mathrm{msg}}$ is the shared message board.  The generated response is projected to a structured team message and environment action.

The marker remains in the token sequence and defines role-turn segments.  If $u_q^-$ is the start of marker $q$, then each token $u$ inherits the latest preceding role:

\begin{equation}
\begin{aligned}
\operatorname{seg}(u)&=\max\{q:u_q^-\leq u\},\\
z(u)&=z_{\operatorname{seg}(u)},\\
\Gamma(u)&=\Gamma_{\operatorname{seg}(u)}.
\end{aligned}
\end{equation}

This construction makes the role used in the prompt identical to the role consumed by the model-side router.  The exact resolver, marker matching, and cached-decoding rules are deferred to the appendix.

\begin{figure*}[t]
\centering
\includegraphics[width=0.9\textwidth]{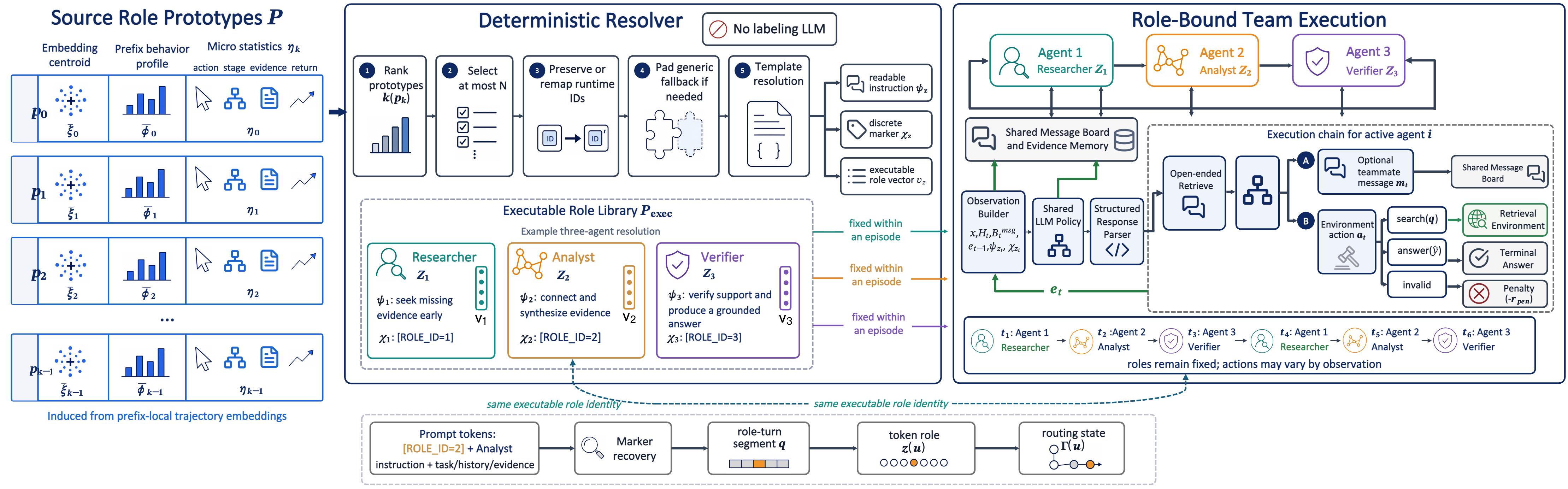}
\caption{Trajectory-to-role binding and execution in ExRole. A deterministic resolver converts trajectory prototypes into executable roles with aligned instructions, markers, and features; fixed role identities then guide round-robin team interaction and model-side sparse routing through shared memory.}
\label{fig:induced-role-profiles}
\end{figure*}

\subsection{Role-Conditioned Sparse Routing}

Prompt conditioning alone does not determine which trainable capacity supports a role.  ExRole-Routed therefore modulates the shared LoRA update at the rank-slot level.  For adapted module $\ell$ and role-turn segment $q$,

\begin{equation}
\mathrm{LoRA}_{\ell}^{\mathrm{role}}(h)
=
B_{\ell}
\left(
\boldsymbol\gamma_{q,\ell}\odot A_{\ell}h
\right)
\alpha_{\ell},
\qquad
\boldsymbol\gamma_{q,\ell}\in\mathbf R^d.
\end{equation}

The router combines the executable-role feature $\mathbf v_{z_q}$, its discrete identity, and a pooled semantic prefix $\zeta_q$:

\begin{equation}
\begin{aligned}
\zeta_q
&=\mathrm{Norm}
\left(
\frac{1}{|\mathcal W_q|}
\sum_{u\in\mathcal W_q}
\mathrm{stopgrad}(\mathcal E(\mathrm{id}_u))
\right),\\
h_q^b
&=F_{\theta_b}(\mathbf v_{z_q})
+E_{\theta_b}(z_q)
+\lambda_c C_{\theta_b}(\zeta_q),\\
\omega_{q,\ell s}
&=\mathrm{Head}^{\mathrm{slot}}_{\theta_b}(h_q^b)_{\ell s}
+\lambda_u\tanh
\left(
\mathrm{Head}^{\mathrm{util}}_{\theta_b}(h_q^b)_{\ell s}
\right),\\
\widehat c_q
&=\mathrm{Head}^{\mathrm{cred}}_{\theta_b}(h_q^b).
\end{aligned}
\end{equation}

Here $\mathcal W_q$ is the prefix window before marker $q$, $\mathrm{id}_u$ is the token identifier (ID) at position $u$, and $\omega_{q,\ell s}$ scores rank slot $s$ of module $\ell$.  The scores define a probability distribution $\varpi_q$ and a hard budget of $S$ directly selected slots:

\begin{equation}
\begin{aligned}
\varpi_{q,\ell s}
&=\mathrm{softmax}(\omega_q/\tau_b)_{\ell s},\\
M_{q,\ell s}
&=\mathbf 1[(\ell,s)\in\mathrm{TopS}(\omega_q)],\\
\delta^{\mathrm{sel}}_{q,\ell s}
&=M_{q,\ell s}(Ld\,\varpi_{q,\ell s}-1),\\
\bar\delta_{q,\ell s}
&=\delta^{\mathrm{sel}}_{q,\ell s}
-\frac{1}{Ld}\sum_{\ell',s'}
\delta^{\mathrm{sel}}_{q,\ell's'},\\
\gamma_{q,\ell s}
&=\mathrm{clip}
\left(
1+\lambda_g\bar\delta_{q,\ell s},
\gamma_{\min},\gamma_{\max}
\right).
\end{aligned}
\end{equation}

The hard mask concentrates role-specific deviations, while centering preserves a balanced shared update before clipping.  Both variants retain the same LoRA parameters and role-conditioned interaction; they differ only in the model-side routing state:

\begin{equation}
\Gamma_q=
\begin{cases}
\mathbf 1, & \text{ExRole-Shared},\\
\{\gamma_{q,\ell s}\}_{\ell,s}, & \text{ExRole-Routed}.
\end{cases}
\end{equation}

\subsection{Turn-Aligned Optimization}

An episode-level reward alone assigns the same credit to every response in a team trajectory.  ExRole retains the group-relative trajectory advantage $\mathrm{Adv}^{(g)}$ from GRPO, while adding discounted local credit to the response tokens produced at turn $t$:

\begin{equation}
\begin{aligned}
U_t^{\mathrm{turn},(g)}
&=\sum_{t'=t}^{T_{\tau^{(g)}}}
\lambda_{\mathrm{disc}}^{t'-t}r_{t'}^{(g)},\\
\widetilde{\mathrm{Adv}}_{t,j}^{(g)}
&=I_{t,j}^{\mathrm{resp},(g)}
\left[
\mathrm{Adv}^{(g)}
+\alpha_c\widehat U_t^{(g)}
\right],
\end{aligned}
\end{equation}

\noindent where $\widehat U_t^{(g)}$ is the normalized turn return and $I_{t,j}^{\mathrm{resp},(g)}$ masks non-response tokens.  The same return supplies a detached role-turn credit target $c_q$ for segment $q$:

\begin{equation}
c_q
=
\mathrm{stopgrad}
\left[
\mathrm{clip}
\left(
\mathcal N_{\mathcal B_R}
(U_{t(q)}^{\mathrm{turn},(g(q))}),-2,2
\right)
\right].
\end{equation}

Here $\mathcal B_R$ is the set of role-turn segments in the current normalization batch.  The router learns to predict this target and associate positive role-turn credit with its selected capacity.  Writing $\tilde c_q=\mathrm{clip}((c_q+2)/4,0,1)$ and letting $\bar\nu_q^{\mathrm{util}}$ denote mean predicted utility over selected slots, the two credit-bearing objectives are

\begin{equation}
\begin{aligned}
\mathcal L_{\mathrm{future}}
&=\frac{1}{2Q}\sum_{q=1}^{Q}
\left[
(\bar\nu_q^{\mathrm{util}}-\tilde c_q)^2
+(\operatorname{sigm}(\widehat c_q)-\tilde c_q)^2
\right],\\
\mathcal L_{\mathrm{credit}}
&=-\frac{1}{Q}\sum_{q=1}^{Q}c_q
\frac{
\sum_{\ell,s}M_{q,\ell s}\log\varpi_{q,\ell s}
}{
\sum_{\ell,s}M_{q,\ell s}
}.
\end{aligned}
\end{equation}

Here $\operatorname{sigm}(\cdot)$ denotes the logistic sigmoid.

Load-balance, role-diversity, sparsity, and entropy terms regularize the remaining routing distribution, while Kullback--Leibler (KL) regularization anchors the policy.  The two model variants are therefore optimized under the same turn-aligned policy objective and differ only by the router loss:

\begin{equation}
\begin{aligned}
\mathcal L_{\mathrm{shared}}
&=-\mathcal J_{\mathrm{GRPO}}^{\mathrm{turn}}
+\beta_{\mathrm{KL}}\mathcal L_{\mathrm{KL}},\\
\mathcal L_{\mathrm{router}}
&=\lambda_{\mathrm{future}}\mathcal L_{\mathrm{future}}
+\lambda_{\mathrm{credit}}\mathcal L_{\mathrm{credit}}
+\mathcal L_{\mathrm{reg}},\\
\mathcal L_{\mathrm{routed}}
&=\mathcal L_{\mathrm{shared}}+\mathcal L_{\mathrm{router}}.
\end{aligned}
\end{equation}

This matched formulation isolates the contribution of role-conditioned sparse LoRA routing: ExRole-Shared and ExRole-Routed use the same role library, interaction protocol, reward, and turn-aligned policy credit.  The appendix gives the full GRPO objective, normalization fallbacks, router regularizers, pseudocode, and hyperparameters.

\begingroup
\section{Experiments}

\subsection{Experimental Setup}

\noindent\textbf{Benchmarks.}
We use MuSiQue and 2WikiMultiHopQA, which require multi-step evidence composition over supporting-document collections.

\noindent\textbf{Baselines.}
The controlled comparison includes single-agent search, multi-agent systems (MAS) without roles or with hand-written roles, and random or shuffled role controls.  All controlled systems share the same policy backbone, action interface, and evaluation protocol.  ExRole-Shared and ExRole-Routed additionally share the induced role library and training objective, differing only in whether LoRA capacity is uniformly shared or role routed.  Table~\ref{tab:external-reference} provides RAG, Search-RL, and MAS-RL references; its base icons distinguish SFT and search-specialized backbones.

\noindent\textbf{Metrics.}
We report \textbf{\emph{Exact Match (EM)}}, \textbf{\emph{token-level F1}}, and \textbf{\emph{strict success (Succ)}}.  EM uses benchmark answer normalization, F1 measures token overlap, and Succ requires a normalized exact answer without substring credit.  Complete data, retrieval, optimization, and sequence-budget settings are given in Supplementary Appendix~B.6; formal metric definitions appear in Supplementary Appendix~C.2.

\subsection{Main Results}

\begin{table*}[t]
\centering
\footnotesize
\newcommand{\mainneg}[2]{\mbox{#1\,\textcolor{bdred}{(#2)}}}
\setlength{\tabcolsep}{2.4pt}
\begin{tabularx}{\textwidth}{@{}Xccccccccc@{}}
\toprule
\textbf{Method} & \textbf{\#Ag.} & \textbf{Role} & \textbf{Cap.} &
\multicolumn{3}{c}{\textbf{MuSiQue}} &
\multicolumn{3}{c}{\textbf{2Wiki}} \\
\cmidrule(lr){5-7}\cmidrule(lr){8-10}
& & & &
\textbf{EM \metricup} & \textbf{F1 \metricup} & \textbf{Succ. \metricup} &
\textbf{EM \metricup} & \textbf{F1 \metricup} & \textbf{Succ. \metricup} \\
\midrule
\multicolumn{10}{@{}l}{\textit{Trajectory-induced roles}} \\
\rowcolor{bdexroleGray}
ExRole-Shared & 3 & Induced & Shared &
\bestplain{31.5} & \bestplain{43.2} & \bestplain{31.5} &
49.0 & 59.1 & 49.0 \\
\rowcolor{bdexroleGray}
ExRole-Routed & 3 & Induced & Routed &
30.0 & 41.5 & 30.0 &
\bestplain{50.0} & \bestplain{59.7} & \bestplain{50.0} \\
\midrule
\multicolumn{10}{@{}l}{\textit{Single-agent baseline}} \\
Single-agent search & 1 & None & Shared &
\mainneg{16.5}{-15.0} & \mainneg{28.8}{-14.4} & \mainneg{16.5}{-15.0} &
\mainneg{36.5}{-13.5} & \mainneg{43.6}{-16.1} & \mainneg{36.5}{-13.5} \\
\midrule
\multicolumn{10}{@{}l}{\textit{Three-agent baselines}} \\
No-role MAS & 3 & None & Shared &
\mainneg{20.0}{-11.5} & \mainneg{31.6}{-11.6} & \mainneg{20.0}{-11.5} &
\mainneg{38.0}{-12.0} & \mainneg{44.5}{-15.2} & \mainneg{38.0}{-12.0} \\
Manual-role MAS & 3 & Human & Shared &
\mainneg{13.0}{-18.5} & \mainneg{23.4}{-19.8} & \mainneg{12.5}{-19.0} &
\mainneg{31.5}{-18.5} & \mainneg{37.8}{-21.9} & \mainneg{31.5}{-18.5} \\
\midrule
\multicolumn{10}{@{}l}{\textit{Role-source controls}} \\
Random role prompt & 3 & Random & Shared &
\mainneg{16.5}{-15.0} & \mainneg{28.0}{-15.2} & \mainneg{16.5}{-15.0} &
\mainneg{36.0}{-14.0} & \mainneg{43.1}{-16.6} & \mainneg{36.0}{-14.0} \\
Shuffled induced role & 3 & Mismatch & Shared &
\mainneg{8.5}{-23.0} & \mainneg{19.1}{-24.1} & \mainneg{8.5}{-23.0} &
\mainneg{42.3}{-7.7} & \mainneg{50.0}{-9.7} & \mainneg{42.3}{-7.7} \\
\bottomrule
\end{tabularx}
\caption{Primary ExRole comparison on MuSiQue and 2WikiMultiHopQA. \#Ag. is the number of agents, Cap. is the trainable-capacity path, and Succ. is strict success. Parenthesized values for controls are differences from the stronger ExRole variant in each metric. The 2Wiki shuffled-role result is averaged over three role-assignment seeds.}
\label{tab:main-controlled-comparison}
\end{table*}

\begin{figure}[!t]
\centering
\begin{minipage}[t]{0.485\linewidth}
  \centering\vspace{0pt}
  \includegraphics[width=0.9\linewidth]{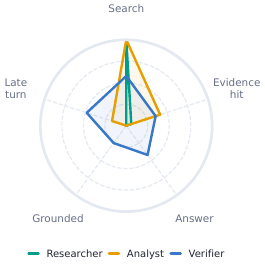}
  {\small\textbf{(a)} Frozen discovery profiles.\par}
\end{minipage}\hfill
\begin{minipage}[t]{0.485\linewidth}
  \centering\vspace{0pt}
  \includegraphics[width=0.9\linewidth]{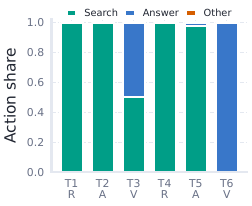}
  {\small\textbf{(b)} Action shares by scheduled turn.\par}
\end{minipage}
\vspace{1.2mm}

\begin{minipage}[t]{0.485\linewidth}
  \centering\vspace{0pt}
  \includegraphics[width=0.9\linewidth]{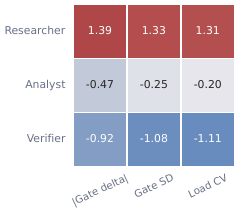}
  {\small\textbf{(c)} Standardized router signatures.\par}
\end{minipage}\hfill
\begin{minipage}[t]{0.485\linewidth}
  \centering\vspace{0pt}
  \includegraphics[width=0.9\linewidth]{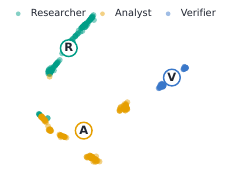}
  {\small\textbf{(d)} Frozen trajectory embeddings.\par}
\end{minipage}
\caption{Role induction and routed execution on MuSiQue. Panels (b) and (c) summarize 200 ExRole-Routed trajectories across scheduled team turns; panel (d) shows a t-SNE projection of frozen role embeddings.}
\label{fig:role-mechanism}
\end{figure}

\noindent\textbf{Controlled role comparison.}
Table~\ref{tab:main-controlled-comparison} compares role sources and trainable capacity under a common evaluation interface.  On MuSiQue, ExRole-Shared obtains 31.5 EM and 43.2 F1, compared with 16.5 EM and 28.8 F1 for single-agent search and 20.0 EM and 31.6 F1 for the no-role MAS.  On 2WikiMultiHopQA, ExRole-Shared and ExRole-Routed reach 49.0/59.1 and 50.0/59.7 EM/F1, respectively.  Both capacity paths preserve the gains from trajectory-induced role conditioning across the two primary benchmarks.

\noindent\textbf{Inference behavior.}
ExRole-Shared and ExRole-Routed use nearly identical numbers of team turns, search calls, retrieved characters, and repeated queries.  Both variants retrieve target-bearing evidence for approximately 73\% of MuSiQue examples and produce grounded answers for 87\%, while strict success remains near 30\%.  This gap localizes much of the remaining error to answer selection and exact answer formation.  Supplementary Appendix~D.5 consolidates the matched cost distributions and stage-wise diagnostics.

\noindent\textbf{Role mechanism.}
Figure~\ref{fig:role-mechanism} connects trajectory-level role induction to routed execution on MuSiQue.  The induced Researcher, Analyst, and Verifier differ in search, evidence, answer, and timing statistics, and their trajectory embeddings form separated regions.  Their scheduled turns also produce distinct action and routing profiles.  The figure characterizes role-conditioned execution under the fixed speaker schedule.

\begin{table}[t]
\centering
\footnotesize
\renewcommand{\arraystretch}{1.14}
\setlength{\tabcolsep}{3pt}
\begin{tabularx}{\columnwidth}{@{}Xccc@{}}
\toprule
\textbf{Method} & \textbf{Base} & \textbf{MuSiQue} &
\textbf{2Wiki} \\
& & \emph{EM / F1} & \emph{EM / F1} \\
\midrule
\rowcolor{bdtealStripe}
\multicolumn{4}{@{}l}{\textit{Retrieval baselines}} \\
Direct & \basesft & 3.0\,/\,11.4 & 22.5\,/\,26.2 \\
Naive RAG & \basesft & 12.5\,/\,22.3 & 26.5\,/\,29.6 \\
\midrule
\rowcolor{bdtealStripe}
\multicolumn{4}{@{}l}{\textit{Search-RL baselines}} \\
Search-R1 & \baseqwen & 33.0\,/\,44.3 & 52.0\,/\,60.3 \\
R1-Searcher & \baseragrl & 47.5\,/\,59.0 & 64.5\,/\,71.2 \\
\midrule
\rowcolor{bdtealStripe}
\multicolumn{4}{@{}l}{\textit{MAS-RL baselines}} \\
MAGRPO & \basesft & 12.0\,/\,23.1 & 26.5\,/\,35.2 \\
Dr. MAS & \basesft & 30.5\,/\,37.8 & 60.0\,/\,67.5 \\
MATPO & \basesft & 36.5\,/\,48.8 & 45.3\,/\,54.4 \\
\midrule
\rowcolor{bdtealStripe}
\multicolumn{4}{@{}l}{\textit{ExRole}} \\
\rowcolor{bdexroleGray}
ExRole-Shared & \basesft & 31.5\,/\,43.2 & 49.0\,/\,59.1 \\
\rowcolor{bdexroleGray}
ExRole-Routed & \basesft & 30.0\,/\,41.5 & 50.0\,/\,59.7 \\
\bottomrule
\end{tabularx}
\caption{External EM/F1 references on the primary benchmarks under each method's stated protocol. Base icons denote \basesft{} shared SFT, \baseqwen{} Search-R1/Qwen, and \baseragrl{} R1-Searcher backbones.}
\label{tab:external-reference}
\end{table}

\begingroup
\begin{figure*}[!t]
\centering
{\footnotesize
\textcolor{bdblue}{\rule[0.5ex]{7pt}{1.1pt}}~MuSiQue\quad
\textcolor{bdgreen}{\rule[0.5ex]{7pt}{1.1pt}}~2Wiki
\par}
\vspace{0.6mm}
\begin{minipage}[t]{0.32\linewidth}
  \centering\vspace{0pt}
  \includegraphics[width=\linewidth]{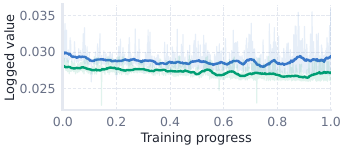}
  {\small\textbf{(a)} Router auxiliary loss.\par}
\end{minipage}\hfill
\begin{minipage}[t]{0.32\linewidth}
  \centering\vspace{0pt}
  \includegraphics[width=\linewidth]{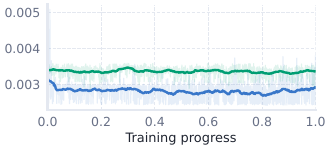}
  {\small\textbf{(b)} Future-utility loss.\par}
\end{minipage}\hfill
\begin{minipage}[t]{0.32\linewidth}
  \centering\vspace{0pt}
  \includegraphics[width=\linewidth]{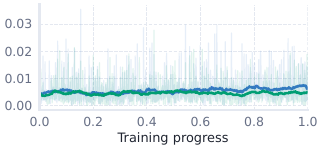}
  {\small\textbf{(c)} Absolute turn credit.\par}
\end{minipage}
\caption{Routed training signals across MuSiQue and 2WikiMultiHopQA. Pale traces are rank-averaged observations; dark traces are centered rolling means over normalized training progress.}
\label{fig:router-training-signals}
\vspace{2.0mm}

{\footnotesize
\textcolor{bdblue}{\rule[0.5ex]{7pt}{1.1pt}}~Shared\quad
\textcolor{bdgreen}{\rule[0.5ex]{7pt}{1.1pt}}~Routed
\par}
\vspace{0.5mm}
\begin{minipage}[t]{0.485\linewidth}
  \centering\vspace{0pt}\includegraphics[width=\linewidth]{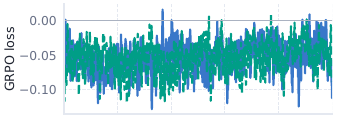}
  {\small\textbf{(a)} MuSiQue policy loss.\par}
\end{minipage}\hfill
\begin{minipage}[t]{0.485\linewidth}
  \centering\vspace{0pt}\includegraphics[width=\linewidth]{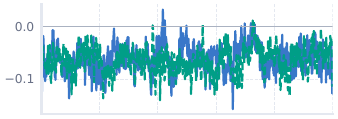}
  {\small\textbf{(b)} 2Wiki policy loss.\par}
\end{minipage}
\vspace{0.9mm}

\begin{minipage}[t]{0.485\linewidth}
  \centering\vspace{0pt}\includegraphics[width=\linewidth]{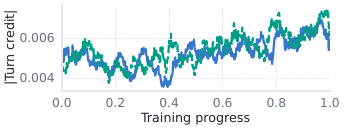}
  {\small\textbf{(c)} MuSiQue turn credit.\par}
\end{minipage}\hfill
\begin{minipage}[t]{0.485\linewidth}
  \centering\vspace{0pt}\includegraphics[width=\linewidth]{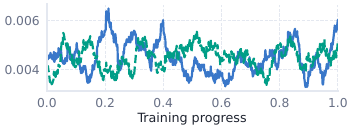}
  {\small\textbf{(d)} 2Wiki turn credit.\par}
\end{minipage}
\caption{Matched Shared and Routed optimization dynamics. The upper row reports GRPO policy loss and the lower row reports absolute turn credit; all curves are rank-averaged centered rolling means over the same 128-update horizon.}
\label{fig:matched-optimization}
\vspace{2.0mm}

\includegraphics[width=\linewidth]{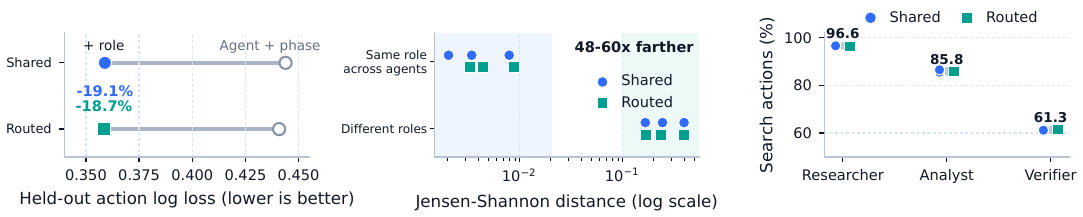}
\begin{minipage}[t]{0.32\linewidth}
  \centering{\small\textbf{(a)} Incremental role information.\par}
\end{minipage}\hfill
\begin{minipage}[t]{0.32\linewidth}
  \centering{\small\textbf{(b)} Cross-agent behavioral consistency.\par}
\end{minipage}\hfill
\begin{minipage}[t]{0.32\linewidth}
  \centering{\small\textbf{(c)} Role-specific action profiles.\par}
\end{minipage}
\caption{Role-Agent-Turn disentanglement on MuSiQue using fixed Shared and Routed checkpoints without additional policy training.}
\label{fig:role-agent-turn-core}
\end{figure*}

\endgroup

\subsection{Role-Agent-Turn Disentanglement}

Using fixed checkpoints, we orthogonally vary speaker order and agent-role assignment on MuSiQue.  Figure~\ref{fig:role-agent-turn-core}(a) shows that role identity reduces held-out action log loss by 19.1\% for Shared and 18.7\% for Routed after accounting for agent identity and turn phase.  Between-role Jensen--Shannon distance is 48--60 times the within-role cross-agent distance, while Researcher, Analyst, and Verifier search rates are 96.6\%, 85.8\%, and 61.3\%.  These results indicate transferable action tendencies rather than fixed agent or turn labels; Supplementary Appendix~D.3 reports the full matrix.

\noindent\textbf{External reference.}
Table~\ref{tab:external-reference} reports representative retrieval, Search-RL, MAS-RL, and ExRole results on the primary benchmarks.  ExRole-Shared reaches 49.0/59.1 EM/F1 on 2WikiMultiHopQA and 31.5/43.2 on MuSiQue; ExRole-Routed reaches 50.0/59.7 and 30.0/41.5.  ExRole is competitive with several MAS-RL systems but remains below the strongest search-specialized policies.

\subsection{Capacity-Path Analysis}

\noindent\textbf{Paired capacity-path comparison.}
We evaluate ExRole-Shared and ExRole-Routed on the same 200 examples for each benchmark.  Both capacity paths preserve the induced-role gains over the corresponding single-agent and no-role controls, while their relative ordering varies across benchmarks.  Supplementary Appendix~D.1 reports the complete paired differences, confidence intervals, and per-example comparisons.

\noindent\textbf{Router training signals.}
Figure~\ref{fig:router-training-signals} reports the routing objectives and turn credit across the two primary benchmarks.  The auxiliary and future-utility losses remain active throughout training, while turn credit varies with sampled trajectories.  Supplementary Appendix~D.1 reports the corresponding task-level effects.

\noindent\textbf{Matched optimization dynamics.}
Figure~\ref{fig:matched-optimization} compares Shared and Routed under the same training horizon.  Its columns correspond to MuSiQue and 2WikiMultiHopQA, while the two rows show smoothed GRPO policy loss and the magnitude of turn-aligned credit.  Both variants receive turn-level credit throughout optimization; only Routed additionally receives the router losses in Figure~\ref{fig:router-training-signals}.

\enlargethispage{2\baselineskip}
\section{Conclusion}

ExRole turns team trajectories into executable roles for interaction and turn-aligned optimization.  On MuSiQue and 2WikiMultiHopQA, induced roles yield distinct search, synthesis, and verification behaviors and outperform single-agent and role-free controls.  Role-Agent-Turn interventions show that these behaviors generalize beyond fixed agents and turn positions, supporting trajectory-induced role binding as a practical framework for structured language-agent collaboration.  This interface makes learned specialization explicit and reusable during policy optimization.

\endgroup
\bibliography{references}

\clearpage
\appendix

\section{Appendix Guide}

\noindent
\begin{tabular}{@{}p{0.15\columnwidth}p{0.76\columnwidth}@{}}
\toprule
Index & Content \\
\midrule
\multicolumn{2}{@{}l}{\textbf{B\quad Method and Experimental Details}} \\
B.1 & \appjump{app.reward-definition}{Complete Environment Objective} \\
B.2 & \appjump{app.role-induction-details}{Role Features, Targets, and Prototype Construction} \\
B.3 & \appjump{app.role-resolution-details}{Deterministic Role Resolution and Marker Alignment} \\
B.4 & \appjump{app.router-details}{Router Objectives and Implementation Details} \\
B.5 & \appjump{app.full-turn-grpo}{Full Turn-Aligned GRPO Objective} \\
B.6 & \appjump{app.experimental-configuration}{Experimental Configuration} \\
B.7 & \appjump{app.manual-role-control-prompt}{Manual-role Control Prompt} \\
B.8 & \appjump{app.role-library-provenance}{Role-Library Provenance and Split Isolation} \\
\addlinespace[2pt]
\multicolumn{2}{@{}l}{\textbf{C\quad Evaluation, Inference, and Optimization Details}} \\
C.1 & \appjump{app.limitations}{Limitations} \\
C.2 & \appjump{app.metric-definitions}{Evaluation Metrics and Agreement Analysis} \\
C.3 & \appjump{app.inference-procedure}{Inference Procedure} \\
C.4 & \appjump{app.sparse-delta-gradient}{Gradient Flow through Balanced Sparse-Delta Routing} \\
C.5 & \appjump{app.training-credit-flow}{Training-Time Credit Flow} \\
C.6 & \appjump{app.training-algorithm}{Turn-Aligned Routed GRPO Algorithm} \\
\addlinespace[2pt]
\multicolumn{2}{@{}l}{\textbf{D\quad Additional Experimental Diagnostics}} \\
D.1 & \appjump{app.paired-routing-effects}{Paired Routing Effects} \\
D.2 & \appjump{app.role-routing-diagnostics}{Role Induction and Evaluation Diagnostics} \\
D.3 & \appjump{app.role-agent-turn-disentanglement}{Role-Agent-Turn Disentanglement} \\
D.4 & \appjump{app.role-credit-complementarity}{Role Induction and Turn-Aligned Credit} \\
D.5 & \appjump{app.inference-diagnostics}{Inference Cost and Answer Diagnostics} \\
D.6 & \appjump{app.hotpot-fullwiki-stress}{Full-Wikipedia HotpotQA Stress Test} \\
D.7 & \appjump{app.external-baseline-curves}{External Reference Training Curves} \\
\addlinespace[2pt]
\multicolumn{2}{@{}l}{\textbf{E\quad Notation}} \\
E.1 & \appjump{app.symbol-table}{Symbol Table} \\
\bottomrule
\end{tabular}

\section{Method and Experimental Details}

\subsection{Complete Environment Objective}
\appanchor{app.reward-definition}

The environment exposes the controlled action space

\begin{equation}
\mathcal A=
\{\mathrm{search}(q_{\mathrm{ret}}),
\mathrm{answer}(\hat y),
\mathrm{invalid}\}.
\end{equation}

Before turn $t$, the available team history is $\mathcal H_t=\{a_{t'},m_{t'},e_{t'}\}_{t'<t}$.  Unless the episode terminates early, speakers follow the round-robin schedule $i_t=1+((t-1)\bmod N)$.  The turn reward used by all ExRole variants is

\begin{equation}
\begin{aligned}
r_t&=r_t^{\mathrm{base}}+r_t^+-\mathrm{Pen}_t,\\
r_t^+
&=\lambda_{\mathrm{ans}}r_t^{\mathrm{ans}}
 +\lambda_{\mathrm{sup}}r_t^{\mathrm{sup}}\\
&\quad +\lambda_{\mathrm{nov}}r_t^{\mathrm{nov}}\\
&\quad +\lambda_{\mathrm{ver}}r_t^{\mathrm{ver}},\\
\mathrm{Pen}_t
&=\lambda_{\mathrm{rep}}\mathrm{Pen}_t^{\mathrm{rep}}
 +\lambda_{\mathrm{early}}\mathrm{Pen}_t^{\mathrm{early}}\\
&\quad +\lambda_{\mathrm{bridge}}\mathrm{Pen}_t^{\mathrm{bridge}}\\
&\quad +\lambda_{\mathrm{insuf}}\mathrm{Pen}_t^{\mathrm{insuf}}\\
&\quad +\lambda_{\mathrm{gwrong}}\mathrm{Pen}_t^{\mathrm{gwrong}}\\
&\quad +\lambda_{\mathrm{unsup}}\mathrm{Pen}_t^{\mathrm{unsup}}\\
&\quad +\lambda_{\mathrm{noans}}\mathrm{Pen}_t^{\mathrm{noans}}.
\end{aligned}
\end{equation}

The shared base term covers action format, communication, answer type, and grounding.  The positive terms reward a correct answer, supporting evidence, evidence novelty, and a grounded verifier decision.  The penalties cover repeated evidence, premature answers, intermediate bridge entities, insufficient evidence, grounded but incorrect answers, unsupported answers, and episodes that terminate without an answer.  With $\mathcal Y(y)$ denoting acceptable aliases,

\begin{equation}
r_t^{\mathrm{ans}}
=
\max_{y'\in\mathcal Y(y)}
\mathbf 1_{\mathrm{succ}}(\hat y_t,y').
\end{equation}

The strict indicator compares $n_{\mathrm{succ}}(\hat y_t)$ with $n_{\mathrm{em}}(y')$, where $n_{\mathrm{succ}}=n_{\mathrm{em}}\circ c_{\mathrm{ans}}$ and $c_{\mathrm{ans}}$ removes the structured action wrapper and common answer prefixes.

\subsection{Role Features, Targets, and Prototype Construction}
\appanchor{app.role-induction-details}

The prefix feature uses the following fixed 30-coordinate schema:

\begin{equation}
\begin{aligned}
\phi_t^{\mathrm{obs}}
&=
\left[
\mathbf r_t^{i}\in\mathbf R^{14},
\mathbf c_t^{i}\in\mathbf R^{4},
\mathbf r_t^{\mathrm{team}}\in\mathbf R^{5},
\right.\\[-1mm]
&\hspace{18mm}\left.
p_t,b_t,\operatorname{onehot}_5(a_t)
\right].
\end{aligned}
\end{equation}

In order, $\mathbf r_t^{i}$ contains the active agent's search, answer, invalid, and message rates; mean message and retrieval lengths; mean, early, and late turn-position statistics; mean shared-board size; accumulated reward contribution; and evidence-hit, grounded-answer, and repair rates.  The count vector $\mathbf c_t^{i}$ contains search, answer, message, and turn counts.  The team vector $\mathbf r_t^{\mathrm{team}}$ contains team search, answer, message, tool-use, and evidence-hit rates; $p_t$ and $b_t$ are progress and remaining-budget ratios.  Rates use the number of observed agent or team turns as their denominator, means are zero for empty sets, and missing scalar metadata are set to zero.  For coordinate $j$ over the $J$ discovery records, the encoder input is

\begin{equation}
\begin{aligned}
\widetilde\phi_{t,j}^{\mathrm{obs}}
&=
\frac{\phi_{t,j}^{\mathrm{obs}}-\mu_j}{\sigma_j+10^{-6}},\\
\mu_j&=\frac{1}{J}\sum_{t=1}^{J}\phi_{t,j}^{\mathrm{obs}},
\qquad
\sigma_j^2=\frac{1}{J}\sum_{t=1}^{J}
(\phi_{t,j}^{\mathrm{obs}}-\mu_j)^2.
\end{aligned}
\end{equation}

Every component is computed from the logged prefix ending at turn $t$.  Future evidence and final return are used only as targets.

For the next-action target, $t_i^+=\min\{t'>t:i_{t'}=i_t\}$ is the next turn of the same agent; if it does not exist, the target is \emph{stop}.  Let $h_{t'}^{\mathrm{evd}}$ denote the logged evidence-hit or grounding indicator at turn $t'$.  The three targets are

\begin{equation}
\begin{aligned}
a_t^+&=
\begin{cases}
a_{t_i^+},&t_i^+\text{ exists},\\
\mathrm{stop},&\text{otherwise},
\end{cases}\\
v_t^{\mathrm{evd}}
&=\mathbf 1\!\left[\exists t'>t:
i_{t'}=i_t\land h_{t'}^{\mathrm{evd}}=1\right],\\
R_t&=R(\tau(t),y_{\tau(t)}).
\end{aligned}
\end{equation}

We use $\lambda_{\mathrm{evd}}=1$ and $\lambda_{\mathrm{ret}}=0.5$ in $\mathcal L_{\mathrm{role}}$.

For a configured discovery-seed set $\mathcal S$, K-means uses Euclidean distance, ten initializations per seed, and the following model-selection statistics:

\begin{equation}
\begin{aligned}
\mathbf k_K^{(s)}
&=\operatorname{KMeans}_{10}(\{\xi_t\}_{t=1}^{J};K,s),\\
\overline{\mathrm{Sil}}(K)
&=\frac{1}{|\mathcal S|}\sum_{s\in\mathcal S}
\operatorname{Sil}(\{\xi_t\},\mathbf k_K^{(s)}),\\
\operatorname{Stability}(K)
&=\frac{2}{|\mathcal S|(|\mathcal S|-1)}
\sum_{s<s'}\operatorname{ARI}
(\mathbf k_K^{(s)},\mathbf k_K^{(s')}),\\
K^\star
&=\arg\max_{K\in\mathcal K}
\left[\overline{\mathrm{Sil}}(K)
+\lambda_{\mathrm{stab}}\operatorname{Stability}(K)\right].
\end{aligned}
\end{equation}

Ties in the last line are resolved by the larger $K$, matching the implementation, and the final assignments use the first configured seed $s_0$: $k_t=k_{K^\star,t}^{(s_0)}$.  Clusters with support below $n_0$ are discarded before prototype construction.

For cluster $k$, let $\mathcal I_k=\{t:k_t=k\}$ and $n_k=|\mathcal I_k|$.  Its centroids and descriptive success lift are

\begin{equation}
\begin{aligned}
\bar\xi_k
&=\frac{1}{n_k}\sum_{t\in\mathcal I_k}\xi_t,\\
\bar\phi_k
&=\frac{1}{n_k}\sum_{t\in\mathcal I_k}\phi_t^{\mathrm{obs}},\\
\mathrm{Lift}_k
&=\frac{1}{n_k}\sum_{t\in\mathcal I_k}
Y_{\tau(t)}^{\mathrm{succ}}
-\frac{1}{J}\sum_{t=1}^{J}Y_{\tau(t)}^{\mathrm{succ}}.
\end{aligned}
\end{equation}

The metadata profile $\eta_k$ records dominant next action, evidence-hit statistics, predicted return, future evidence hit, and final return.  Readable instructions are generated without an additional language model.  Each cluster is scored against a fixed functional vocabulary,

\begin{equation}
\begin{aligned}
S_{k,\upsilon}
&=\mathrm{TemplateScore}_{\upsilon}(\bar\phi_k,\eta_k),\\
\mathcal U_{\mathrm{role}}
&=\{\mathrm{Researcher},\mathrm{Coordinator},
\mathrm{Verifier},\mathrm{Analyst}\}.
\end{aligned}
\end{equation}

High early-search and evidence-hit rates favor a research template, whereas late grounded answers favor a verification template.  Let $\mathcal G(S)$ greedily scan all $(k,\upsilon)$ pairs in decreasing $(S_{k,\upsilon},k,\upsilon)$ order, accepting a pair only when neither its cluster nor functional type has been assigned.  The last coordinate follows reverse lexicographic order over the fixed vocabulary, matching the implementation.  Any cluster left unmatched after this uniqueness pass takes its own highest-scoring type, with ties resolved by the vocabulary order Researcher, Coordinator, Verifier, Analyst:

\begin{equation}
\begin{aligned}
\upsilon_k&=
\begin{cases}
\upsilon,&(k,\upsilon)\in\mathcal G(S),\\
\arg\max_{\upsilon\in\mathcal U_{\mathrm{role}}}
S_{k,\upsilon},&k\notin\operatorname{dom}\mathcal G(S),
\end{cases}\\
(\psi_k,\chi_k)
&=\operatorname{Template}_{\upsilon_k}(\bar\phi_k,\eta_k).
\end{aligned}
\end{equation}

Thus the role type of one cluster can depend on the scores of other clusters; the resolver is a joint deterministic map, not a pointwise template lookup.

We set $\lambda_{\mathrm{stab}}=0.2$ in the role-count criterion and discard clusters below the minimum support $n_0$.  MuSiQue fixes $K=3$ to the three-agent execution budget.  HotpotQA and 2WikiMultiHopQA search $K\in\{2,3,4\}$ and select $K=4$ before resolving three executable roles.  The broader sensitivity analysis in the appendix is post-hoc and is not used for model selection.

\subsection{Deterministic Role Resolution and Marker Alignment}
\appanchor{app.role-resolution-details}

The resolver ranks source prototypes using

\begin{equation}
\begin{aligned}
\kappa(p_k)&=
\left(
\mathrm{stage}(p_k),
-\eta_k^{\mathrm{early}},
\eta_k^{\mathrm{late}},
\right.\\[-1mm]
&\hspace{14mm}\left.
\eta_k^{\mathrm{pos}},
-\mathrm{Lift}_k,
-n_k,
k
\right).
\end{aligned}
\end{equation}

The resolver sorts $\kappa$ lexicographically in ascending order, so coordinator and planner roles precede researcher and solver roles, followed by analyst and synthesizer roles and then verifier roles; the final source-cluster ID breaks exact ties.  Let $\mathcal P_N^{\mathrm{sel}}$ be the first $N$ prototypes under this ordering, $K_{\mathrm{sel}}=|\mathcal P_N^{\mathrm{sel}}|$, and $\iota_i$ the source ID of its $i$th entry.  Source IDs are preserved when they already form a contiguous runtime range; otherwise they are remapped by execution position:

\begin{equation}
\begin{aligned}
\mathcal P_N^{\mathrm{sel}}
&=\mathrm{Take}_N(\mathrm{Sort}_{\kappa}(\mathcal P)),\\
z_i
&=
\begin{cases}
\iota_i,
&\{\iota_j\}_{j=1}^{K_{\mathrm{sel}}}
=\{0,\ldots,K_{\mathrm{sel}}-1\},\\
i-1,&\text{otherwise},
\end{cases}\\
\mathcal P_{\mathrm{exec}}
&=\mathrm{Sort}_z
\left(
\mathrm{PadGeneric}
(\{(\mathcal P_{N,i}^{\mathrm{sel}},z_i)\}_{i=1}^{K_{\mathrm{sel}}},N)
\right).
\end{aligned}
\end{equation}

Source IDs remain in metadata for auditability, while the router indexes only contiguous runtime role IDs.  If fewer than $N$ roles survive, the resolver appends a generic collaborative fallback with a fresh marker.

Equivalent tokenizations of each marker are matched, overlapping occurrences are coalesced, and the resulting spans are ordered as

\begin{equation}
\mathcal M=
\{(u_q^-,u_q^+,z_q)\}_{q=1}^{Q},
\qquad u_1^-<\cdots<u_Q^-.
\end{equation}

Tokens before the first marker use the first segment, and cached decoding retains the latest segment ID.  A malformed sequence without a marker falls back to runtime role $0$; formal ExRole prompts always include a marker.

\begin{algorithm}[H]
\caption{Deterministic assignment of induced roles}
\label{alg:role-assignment}
\begin{tcolorbox}[
  colback=bdtealStripe,
  colframe=bdtealLine,
  boxrule=0.45pt,
  arc=2pt,
  left=4pt,
  right=4pt,
  top=3pt,
  bottom=3pt
]
\begingroup
\small
\renewcommand{\algorithmicrequire}{\alginputicon\ \textbf{Input:}}
\renewcommand{\algorithmicensure}{\algoutputicon\ \textbf{Output:}}
\begin{algorithmic}[1]
\REQUIRE Source prototypes $\mathcal P$ and number of agents $N$
\ENSURE Runtime roles $(z_1,\ldots,z_N)$ and marker-conditioned observations
\STATE Score prototypes with $\kappa(p_k)$ and retain the first $N$
\WHILE{fewer than $N$ roles are available}
  \STATE Append a generic collaborative role with a fresh marker
\ENDWHILE
\STATE Preserve contiguous source IDs; otherwise remap by execution position
\STATE Sort executable roles by runtime role ID
\FOR{$i=1$ to $N$}
  \STATE Bind role $z_i$ and marker $\chi_{z_i}$ to agent $i$
\ENDFOR
\RETURN $(z_1,\ldots,z_N)$ and updated observation templates
\end{algorithmic}
\endgroup
\end{tcolorbox}
\end{algorithm}

\subsection{Router Objectives and Implementation Details}
\appanchor{app.router-details}

The main text writes the normalized routing distribution as $\varpi_{q,\ell s}$.  Equivalently, a total soft budget $S$ is distributed across all adapted-module--rank slots:

\begin{equation}
\tilde b_{q,\ell s}
=S\varpi_{q,\ell s},
\qquad
\bar b_q
=\frac{1}{Ld}\sum_{\ell,s}\tilde b_{q,\ell s}
=\frac{S}{Ld}.
\end{equation}

Hence $\tilde b_{q,\ell s}/\bar b_q=Ld\,\varpi_{q,\ell s}$, which yields the compact sparse-delta expression in the main text.  The selected-slot utility prediction is

\begin{equation}
\bar\nu_q^{\mathrm{util}}
=
\frac{
\sum_{\ell,s}M_{q,\ell s}
\operatorname{sigm}(\nu_{q,\ell s}^{\mathrm{util}})
}{
\sum_{\ell,s}M_{q,\ell s}
}.
\end{equation}

Let $\bar\varpi_{\ell s}=Q^{-1}\sum_q\varpi_{q,\ell s}$.  The remaining regularizers are

\begin{equation}
\begin{aligned}
\mathcal L_{\mathrm{load}}
&=\frac{1}{Ld}\sum_{\ell,s}
\left(\bar\varpi_{\ell s}-\frac{1}{Ld}\right)^2,\\
\mathcal L_{\mathrm{div}}
&=\frac{1}{N(N-1)}
\sum_{z\neq z'}
\frac{
\bar{\boldsymbol\varpi}_z^\top\bar{\boldsymbol\varpi}_{z'}
}{
\|\bar{\boldsymbol\varpi}_z\|_2
\|\bar{\boldsymbol\varpi}_{z'}\|_2
},\\
\mathcal L_{\mathrm{l1}}
&=\frac{1}{QLd}\sum_{q,\ell,s}
\tilde b_{q,\ell s}(1-M_{q,\ell s}),\\
\mathcal L_{\mathrm{ent}}
&=\frac{1}{Q}\sum_{q,\ell,s}
\varpi_{q,\ell s}\log\varpi_{q,\ell s},\\
\mathcal L_{\mathrm{reg}}
&=\lambda_{\mathrm{load}}\mathcal L_{\mathrm{load}}
+\lambda_{\mathrm{div}}\mathcal L_{\mathrm{div}}
+\lambda_{\mathrm{l1}}\mathcal L_{\mathrm{l1}}
+\lambda_{\mathrm{ent}}\mathcal L_{\mathrm{ent}}.
\end{aligned}
\end{equation}

Here $\bar{\boldsymbol\varpi}_z$ is obtained by evaluating executable role $z$ under the minibatch-mean semantic context.  This keeps role diversity active even when a microbatch contains only one observed role.  Since $\mathcal L_{\mathrm{ent}}$ is negative entropy, minimizing it discourages premature slot concentration.

\begin{table}[H]
\centering
\footnotesize
\setlength{\tabcolsep}{4pt}
\begin{tabular}{ll}
\toprule
\textbf{Component} & \textbf{Setting} \\
\midrule
LoRA structure & all-linear, rank $d=8$, scale $16$ \\
Adapted modules & $L=196$ \\
Direct slot budget & $S=392$ of $Ld=1568$ slots \\
Router state & hidden size $128$, prefix window $256$ \\
Routing coefficients & $\tau_b=0.8$, $\lambda_u=0.35$, $\lambda_c=0.5$ \\
Gate & $\lambda_g=0.5$, range $[0.5,1.5]$ \\
Router losses & $\lambda_{\mathrm{future}}=0.05$, $\lambda_{\mathrm{credit}}=0.10$ \\
Regularization & $\lambda_{\mathrm{load}}=0.10$, $\lambda_{\mathrm{div}}=0.05$ \\
Small regularizers & $\lambda_{\mathrm{l1}}=\lambda_{\mathrm{ent}}=10^{-4}$ \\
Optimization & $\beta_{\mathrm{KL}}=10^{-3}$, router gradient scale $1024$ \\
\bottomrule
\end{tabular}
\caption{Role-routing architecture and optimization settings.}
\label{tab:router-hyperparameters}
\end{table}

\subsection{Full Turn-Aligned GRPO Objective}
\appanchor{app.full-turn-grpo}

For each task, GRPO samples $G$ trajectories and computes

\begin{equation}
\begin{aligned}
\mu_R
&=\frac{1}{G}\sum_{g=1}^{G}R(\tau^{(g)},y),\\
\sigma_R
&=\sqrt{\frac{1}{G}\sum_{g=1}^{G}
\left(R(\tau^{(g)},y)-\mu_R\right)^2},\\
\mathrm{Adv}^{(g)}
&=\frac{R(\tau^{(g)},y)-\mu_R}{\sigma_R+\epsilon}.
\end{aligned}
\end{equation}

The actor-side turn return is normalized within each trajectory:

\begin{equation}
\widehat U_t^{(g)}
=
\begin{cases}
\dfrac{U_t^{\mathrm{turn},(g)}-\mu_U^{(g)}}{\sigma_U^{(g)}},
&\sigma_U^{(g)}\geq10^{-6},\\[2mm]
\mathrm{clip}(U_t^{\mathrm{turn},(g)}-\mu_U^{(g)},-2,2),
&\text{otherwise}.
\end{cases}
\end{equation}

Thus, a one-turn or constant-return trajectory contributes zero actor-side local credit.  The router normalizer instead retains a raw singleton or zero-variance target before clipping:

\begin{equation}
\mathcal N_{\mathcal B_R}(v_q)
=
\begin{cases}
\dfrac{v_q-\mu_{\mathcal B_R}}{\sigma_{\mathcal B_R}},
&|\mathcal B_R|>1\ \text{and}\ \sigma_{\mathcal B_R}\geq10^{-6},\\[2mm]
v_q,&\text{otherwise},
\end{cases}
\end{equation}

where $v_q=U_{t(q)}^{\mathrm{turn},(g(q))}$.  For response token $j$ at turn $t$, define

\begin{equation}
\begin{aligned}
\rho_{t,j}^{(g)}
&=
\frac{
\pi_{\theta,\theta_b}
(\mathrm{tok}_{t,j}^{(g)}\mid o_t^{(g)},
\mathrm{tok}_{t,<j}^{(g)})
}{
\pi_{\theta_{\mathrm{old}},\theta_{b,\mathrm{old}}}
(\mathrm{tok}_{t,j}^{(g)}\mid o_t^{(g)},
\mathrm{tok}_{t,<j}^{(g)})
},\\
\bar\rho_{t,j}^{(g)}
&=\mathrm{clip}
\left(\rho_{t,j}^{(g)},1-\epsilon_{\mathrm{clip}},1+\epsilon_{\mathrm{clip}}\right).
\end{aligned}
\end{equation}

The turn-aligned policy objective is

\begin{equation}
\mathcal J_{\mathrm{GRPO}}^{\mathrm{turn}}
=
{\bf E}_{x,g,t,j}
\left[
\min
\left(
\rho_{t,j}^{(g)}\widetilde{\mathrm{Adv}}_{t,j}^{(g)},
\bar\rho_{t,j}^{(g)}\widetilde{\mathrm{Adv}}_{t,j}^{(g)}
\right)
\right].
\end{equation}

We use $\lambda_{\mathrm{disc}}=0.9$ and $\alpha_c=0.35$.  Router-only gradients are multiplied by $1024$ after backpropagation and before global clipping; LoRA gradients are not rescaled.  This changes the router's optimization scale but not the forward gate or the objective.

\subsection{Experimental Configuration}
\appanchor{app.experimental-configuration}

The primary MuSiQue and 2WikiMultiHopQA evaluations use the same team-search interface and metric implementation.  ExRole-Shared and ExRole-Routed also use the same benchmark-specific warm start, reward, training horizon, and turn-aligned credit; their only architectural difference is the capacity-routing path.  HotpotQA is reported separately as a full-Wikipedia retrieval stress test.

\begin{table}[H]
\centering
\footnotesize
\setlength{\tabcolsep}{4pt}
\begin{tabular}{@{}L{0.27\columnwidth}L{0.64\columnwidth}@{}}
\toprule
\textbf{Component} & \textbf{Configuration} \\
\midrule
Primary benchmarks & MuSiQue and 2WikiMultiHopQA; 200 held-out examples per benchmark \\
Stress test & HotpotQA with 200 held-out examples and a lexical full-Wikipedia index \\
Retrieval & Supporting-document collections for the primary benchmarks; top-3 results with 640 characters per retrieved result \\
Team interface & Three agents, round-robin execution, at most six team turns, and 640 characters per retrieved result \\
Policy backbone & Qwen2.5-7B SFT with all-linear LoRA, rank 8 and $\alpha=16$ \\
Training data & 512 benchmark-specific instances; batch size 4 \\
Policy optimization & GRPO group size 8, learning rate $10^{-6}$, and 128 policy updates \\
Sequence budgets & Data prompt 4096, rollout prompt 6144, response 8192, and total model context 16384 tokens \\
Role induction & Three source prototypes for MuSiQue; four for 2WikiMultiHopQA and the HotpotQA stress test; three executable roles after resolution \\
Capacity variants & Shared uses a uniform LoRA path; Routed directly selects 392 of 1568 module--rank slots and applies the centered sparse-delta gate \\
\bottomrule
\end{tabular}
\caption{Training and evaluation configuration for the primary benchmarks and the full-Wikipedia stress test. Router-specific coefficients are reported in Table~\ref{tab:router-hyperparameters}.}
\label{tab:experimental-configuration}
\end{table}

\subsection{Manual-role Control Prompt}
\appanchor{app.manual-role-control-prompt}

The \texttt{manual\_role} control uses a fixed, ordered library of planner, solver, and verifier instructions.  Its training configuration sets \texttt{role\_mode: manual\_role} and \texttt{specialization\_mode: none}.  The evaluator resolves the three entries in the displayed order and applies the same per-turn team-search template, state fields, and action parser as ExRole.  The implementation renders one prompt string per turn rather than separate API system messages; the box separates its shared context, manual instruction, and action contract for readability.  Bracketed fields are filled from the common environment state.

\begin{tcolorbox}[
  breakable,
  colback=bdtealStripe,
  colframe=bdtealLine,
  boxrule=0.45pt,
  arc=2pt,
  left=4pt,
  right=4pt,
  top=3pt,
  bottom=3pt,
  title=\textbf{Manual-role control prompt},
  title after break=\textbf{Manual-role control prompt (continued)}
]
\footnotesize
\textbf{System/shared context.} The common prompt begins with
\begin{quote}
\ttfamily
You are part of a \{num\_agents\}-agent team solving a structured search task.\\
Question: \{question\}\\
Current team turn: \{turn\}/\{max\_turns\}\\
Current speaker: Agent \{agent\_id\} (\{role\_name\})\\
Your functional bias: \{role\_prompt\}
\end{quote}
It then inserts the latest retrieved evidence, exact prior search queries, shared message board, and recent team history; it also appends the structured tag \texttt{\detokenize{[ROLE_ID=k]}}, role-specific operating rules, an optional task answer contract, and the common final-turn rule.

\textbf{Manual instruction and name-dispatched rules.} The role IDs follow the listed order.
\renewcommand{\arraystretch}{1.08}
\begin{tabularx}{\linewidth}{@{}p{0.22\linewidth}X@{}}
\texttt{planner} & ``Prioritize task decomposition, decide what evidence is still missing, and send concise coordination messages before issuing searches.''  Additional rules: ``Route the team away from duplicated work and name the next missing evidence gap concretely.''  ``Prefer one short message or one focused search over a broad speculative query.'' \\
\addlinespace[2pt]
\texttt{solver} & ``Focus on targeted evidence gathering and drafting candidate answers from retrieved information.''  Additional rules: ``Synthesize the current evidence into one candidate answer or one narrower follow-up query.''  ``Avoid broad repeat searches; only search again when you can name the concrete missing fact.''  ``If the current evidence already points to one short plausible span, prefer \texttt{\detokenize{<answer>}} over another broad search.'' \\
\addlinespace[2pt]
\texttt{verifier} & ``Check consistency of the candidate answer against the gathered evidence and only finalize when the evidence is sufficient.''  Additional rules: ``Audit candidate facts against the retrieved evidence and prefer \texttt{\detokenize{<answer>}} over another \texttt{\detokenize{<search>}} once one plausible short span appears.''  ``Only search again when you are resolving one explicit contradiction or one missing entity link.''  ``Do not leave the episode unfinished if the evidence already supports a grounded exact span.'' \\
\end{tabularx}

\textbf{Action contract shared with ExRole.} Each response must contain exactly one final environment action:
\begin{quote}
\ttfamily
\detokenize{<search>your query</search>}\\
\detokenize{<answer>your final answer</answer>}
\end{quote}
Before that action, the prompt permits only a very short \texttt{\detokenize{<think>...</think>}} and one short \texttt{\detokenize{<message>...</message>}}.  Missing evidence requires \texttt{\detokenize{<search>}}, sufficient evidence requires \texttt{\detokenize{<answer>}}, and an answer must be only the shortest exact evidence span rather than a sentence or explanation.  Unsupported spans require another search, exact prior queries may be repeated only with a new disambiguating term, and malformed output without exactly one final action is penalized.  On the final team turn, the shared rule requires \texttt{\detokenize{<answer>}} rather than another search; a last-turn search is penalized and ends the episode without a final answer.
\end{tcolorbox}

Thus the manual-role and ExRole conditions share the same team interface, state visibility, role-marker format, and action grammar.  They differ in how the role instruction is obtained: manual-role uses the fixed library above, whereas ExRole derives executable role instructions from trajectories; only the routed ExRole variant additionally changes the capacity path.

\subsection{Role-Library Provenance and Split Isolation}
\appanchor{app.role-library-provenance}

For the MuSiQue role library used in the matched MuSiQue experiments, we conducted a provenance and split-isolation audit before policy optimization and evaluation.  The frozen library was induced only from 64 \texttt{train} bootstrap trajectories, comprising 323 prefix records; the role encoder consumes prefix-local behavioral statistics rather than question text or retrieved document content.  Against the 200-example validation set, we found zero overlap in task identifiers, normalized question strings, answer strings, hop identifiers, or gold supporting-document contents.  Because both splits draw contextual passages from Wikipedia, 57 non-gold page titles and one non-gold raw passage recur across the two collections; the repeated passage does not contain the validation answer and is not a supporting document for the corresponding evaluation item.  The static role artifact is frozen before policy training, loaded only for role assignment at episode initialization, and is not rebuilt or updated from evaluation rollouts.  Future-evidence and return targets are computed solely from the source \texttt{train} trajectories.  Thus, the audit establishes isolation at the task, answer, gold-evidence, and role-induction levels while distinguishing this guarantee from the unavoidable reuse of a common background corpus.

\section{Evaluation, Inference, and Optimization Details}

\subsection{Limitations}
\appanchor{app.limitations}

Our primary evaluation focuses on supporting-evidence multi-hop question answering with a fixed team-search interface.  Generalization to software engineering, web navigation, longer tool-use episodes, and larger teams remains untested.  Appendix~\appjump{app.hotpot-fullwiki-stress}{D.6} separately reports a full-Wikipedia retrieval stress test in which induced roles do not outperform the strongest controls.

ExRole induces its role library offline from prior trajectories and keeps agent-role assignments fixed within an episode.  A substantial shift in the task distribution, retrieval system, or tool interface may therefore require the role library to be refreshed.

The turn-level reward components are manually weighted, and routed capacity does not improve accuracy on every benchmark.  Learned credit models, adaptive role assignment, and larger role libraries are promising directions for more robust specialization.

\subsection{Evaluation Metrics and Agreement Analysis}
\appanchor{app.metric-definitions}

Let $\hat{y}_n$ be the final answer for evaluation example $n\in\{1,\ldots,N_{\mathrm{eval}}\}$ and let $\mathcal{Y}_n$ be its set of acceptable gold answers.  The benchmark normalizer $n_{\mathrm{em}}(\cdot)$ lowercases text, removes punctuation and English articles, and collapses whitespace.  Per-example exact match is

\begin{equation}
\mathrm{EM}_n
=
\max_{y'\in\mathcal{Y}_n}
\mathbf{1}
\!\left[
n_{\mathrm{em}}(\hat{y}_n)=n_{\mathrm{em}}(y')
\right].
\end{equation}

For token-level F1, let $\widehat{\mathcal{T}}_n$ and $\mathcal{T}_n^{y'}$ be the token multisets obtained from $n_{\mathrm{em}}(\hat{y}_n)$ and $n_{\mathrm{em}}(y')$.  Their multiset intersection counts repeated tokens up to the smaller multiplicity.  We compute

\begin{equation}
\begin{aligned}
\mathrm{Ov}_n(y')
&=|\widehat{\mathcal{T}}_n\cap\mathcal{T}_n^{y'}|,\\
\mathrm{Prec}_n(y')
&=\frac{\mathrm{Ov}_n(y')}{\max(1,|\widehat{\mathcal{T}}_n|)},\\
\mathrm{Rec}_n(y')
&=\frac{\mathrm{Ov}_n(y')}{\max(1,|\mathcal{T}_n^{y'}|)},\\
\mathrm{F1}_n &= \max_{y'\in\mathcal{Y}_n}
\frac{2\,\mathrm{Prec}_n(y')\mathrm{Rec}_n(y')}
{\mathrm{Prec}_n(y')+\mathrm{Rec}_n(y')},
\end{aligned}
\end{equation}

where the last fraction is defined as zero when both precision and recall are zero.  Let $c_{\mathrm{ans}}(\cdot)$ remove the structured action wrapper, common answer prefixes, surrounding whitespace, and terminal punctuation.  The environment normalizer is the composition $n_{\mathrm{succ}}=n_{\mathrm{em}}\circ c_{\mathrm{ans}}$.  Strict success compares this cleaned prediction with the benchmark-normalized gold answer, so substring matches receive no credit:

\begin{equation}
\mathrm{Succ}_n
=
\max_{y'\in\mathcal{Y}_n}
\mathbf{1}_{\mathrm{succ}}(\hat{y}_n,y').
\end{equation}

The reported dataset-level scores are percentages,

\begin{equation}
(\mathrm{EM},\mathrm{F1},\mathrm{Succ})
=
\frac{100}{N_{\mathrm{eval}}}
\sum_{n=1}^{N_{\mathrm{eval}}}
(\mathrm{EM}_n,\mathrm{F1}_n,\mathrm{Succ}_n).
\end{equation}

EM and Succ therefore need not be identical because only Succ applies the action-and-prefix cleanup $c_{\mathrm{ans}}$.  We quantify their per-example agreement using

\begin{equation}
D_{\mathrm{Succ,EM}}
=
\sum_{n=1}^{N_{\mathrm{eval}}}
\mathbf{1}[\mathrm{Succ}_n\neq\mathrm{EM}_n].
\end{equation}

For the Shared and Routed evaluations on the two primary 200-example splits, $D_{\mathrm{Succ,EM}}=0$: the independently computed rules happen to agree on those predictions.  The same equality also holds for the two HotpotQA stress-test variants, whereas several HotpotQA control rows have nonzero disagreement because the additional cleanup changes a small number of decisions.  We always compute EM, F1, and Succ independently rather than copying one column into another.

\subsection{Inference Procedure}
\appanchor{app.inference-procedure}

At inference time, role induction is performed before evaluation rather than repeated for every sample.  ExRole uses a role library induced from prior trajectories, assigns one role prompt to each agent, and then executes a team-search episode.  The procedure is:

\begin{enumerate}
\item Load source prototypes ${\cal P}=\{p_k\}_{k=0}^{K-1}$ from the induced role library.
\item Select and normalize the executable set ${\cal P}_{\mathrm{exec}}$, resolve agent-level runtime roles $z_1,\ldots,z_N$, and insert marker $\chi_{z_i}$ into each agent observation.
\item At turn $t$, choose the active speaker by the team schedule, usually $i_t=1+((t-1)\bmod N)$.
\item Build the observation $o_t$ from the task, active role, latest evidence, shared board, recent search queries, and recent team history.
\item During prefill, coalesce every role marker in the multi-turn sequence and align each token to its role-turn segment; during cached decoding, retain the latest segment ID.
\item Pool the semantic prefix for the active segment and compute slot scores $\omega_{q,\ell s}$, soft budget $\tilde{b}_{q,\ell s}$, top-$S$ mask $M_{q,\ell s}$, and balanced sparse-delta multiplier $\gamma_{q,\ell s}$.
\item Generate the assistant response with the role-conditioned LoRA path.
\item Project the response to a structured action: search, answer, or invalid.
\item If the action is search, update evidence and continue.  If the action is answer, terminate and score the answer.
\end{enumerate}

\subsection{Gradient Flow through Balanced Sparse-Delta Routing}
\appanchor{app.sparse-delta-gradient}

The forward pass treats the top-$S$ mask as fixed while retaining a differentiable soft budget on the selected slots.  Define

\begin{equation}
\begin{aligned}
\delta^{\mathrm{sel}}_{q,\ell s}
&=M_{q,\ell s}
\left(\frac{\tilde b_{q,\ell s}}{\bar b_q}-1\right),\\
\bar\delta_{q,\ell s}
&=\delta^{\mathrm{sel}}_{q,\ell s}
-\frac{1}{Ld}\sum_{\ell',s'}\delta^{\mathrm{sel}}_{q,\ell's'},\\
\gamma_{q,\ell s}
&=\operatorname{clip}\!\left(1+\lambda_g\bar\delta_{q,\ell s},
\gamma_{\min},\gamma_{\max}\right).
\end{aligned}
\end{equation}

Away from clipping boundaries and while the discrete mask is unchanged, the softmax couples every routing score to every soft budget.  The exact chain rule is

\begin{equation}
\frac{\partial \gamma_{q,\ell s}}
{\partial \omega_{q,\ell' s'}}
=
\lambda_g
\sum_{\ell''=1}^{L}\sum_{s''=1}^{d}
\frac{\partial \bar\delta_{q,\ell s}}
{\partial \tilde b_{q,\ell''s''}}
\frac{\partial \tilde b_{q,\ell''s''}}
{\partial \omega_{q,\ell' s'}}.
\end{equation}

Selected slots receive direct role-specific deviations, while centering propagates a compensating shift to all slots.  The future-utility, role-turn-credit, load-balance, diversity, sparsity, and entropy objectives additionally act directly on the soft routing distribution.  Consequently, router learning does not rely on differentiating through the top-$S$ membership itself.

\subsection{Training-Time Credit Flow}
\appanchor{app.training-credit-flow}

For a rollout group, GRPO first converts episode rewards into normalized trajectory advantages.  In parallel, ExRole discounts the recorded turn rewards and aligns each return to the role-turn segment that produced it:

\begin{equation}
\begin{aligned}
x
&\rightarrow \{\tau^{(g)}\}_{g=1}^{G},\\
\{\tau^{(g)}\}_{g=1}^{G}
&\rightarrow \{r_t^{(g)}\}_{g,t}
\rightarrow \{U_t^{\mathrm{turn},(g)}\}_{g,t},\\
\left(R(\tau^{(g)},y),U_t^{\mathrm{turn},(g)}\right)
&\rightarrow \widetilde{\mathrm{Adv}}_{t,j}^{(g)}
\rightarrow \mathcal J_{\mathrm{GRPO}}^{\mathrm{turn}},\\
U_t^{\mathrm{turn},(g)}
&\rightarrow
\left({\cal L}_{\mathrm{future}},
{\cal L}_{\mathrm{credit}}\right),
\end{aligned}
\end{equation}

The actor keeps the group-relative trajectory advantage and adds a weighted, normalized turn return only to the responsible role-turn segment.  The router uses the same detached turn return as its future-utility and role-turn credit target.  Thus a high-reward role turn reinforces both its generated action and its selected capacity path, while global role diversity, load balance, entropy, and sparsity regularizers prevent routing collapse.

\subsection{Turn-Aligned Routed GRPO Algorithm}
\appanchor{app.training-algorithm}

Algorithm~\ref{alg:turn-aligned-routed-grpo} summarizes the complete update procedure corresponding to the objectives in the main text.

\begin{algorithm}[H]
\caption{Turn-aligned routed GRPO update}
\label{alg:turn-aligned-routed-grpo}
\begin{tcolorbox}[
  colback=bdtealStripe,
  colframe=bdtealLine,
  boxrule=0.45pt,
  arc=2pt,
  left=4pt,
  right=4pt,
  top=3pt,
  bottom=3pt
]
\begingroup
\small
\renewcommand{\algorithmicrequire}{\alginputicon\ \textbf{Input:}}
\renewcommand{\algorithmicensure}{\algoutputicon\ \textbf{Output:}}
\begin{algorithmic}[1]
\REQUIRE Task $x$, gold answer $y$, role library $\mathcal{P}$, policy $\theta$, and router $\theta_b$
\ENSURE Updated policy $\theta$ and router $\theta_b$
\STATE \algstageicon\ \textbf{Roll out teams.} Sample $\{\tau^{(g)}\}_{g=1}^{G}$ and record turn rewards $\{r_t^{(g)}\}$
\STATE \algstageicon\ \textbf{Compute group credit.} Normalize episode rewards into $\mathrm{Adv}^{(g)}$
\FOR{each trajectory $\tau^{(g)}$}
  \STATE \algstageicon\ \textbf{Align role turns.} Map every response token to its role-turn segment
  \STATE \algstageicon\ \textbf{Discount local credit.} Compute $U_t^{\mathrm{turn},(g)}$ and $\widehat{U}_t^{(g)}$
  \STATE \algstageicon\ \textbf{Route capacity.} Pool $\zeta_q$ and compute balanced gate $\gamma_{q,\ell s}$
  \STATE \algstageicon\ \textbf{Align policy credit.} Form token advantages $\widetilde{\mathrm{Adv}}_{t,j}^{(g)}$
\ENDFOR
  \STATE \algstageicon\ \textbf{Optimize.} Compute $\mathcal J_{\mathrm{GRPO}}^{\mathrm{turn}}$ and $\mathcal L_{\mathrm{router}}$
\STATE Form $\mathcal L_{\mathrm{routed}}=\mathcal L_{\mathrm{shared}}+\mathcal L_{\mathrm{router}}$ and backpropagate
  \STATE Multiply router-only gradients by $1024$, apply global gradient clipping, and update $(\theta,\theta_b)$
\RETURN \algoutputicon\ Updated $(\theta,\theta_b)$
\end{algorithmic}
\endgroup
\end{tcolorbox}
\end{algorithm}

\section{Additional Experimental Diagnostics}

\subsection{Paired Routing Effects}
\appanchor{app.paired-routing-effects}

\begin{figure}[H]
\centering
\includegraphics[width=\columnwidth]{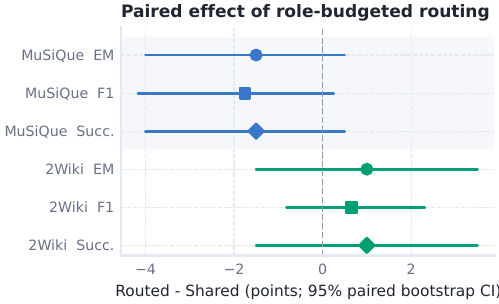}
\caption{Paired effect of role-conditioned sparse LoRA routing. Points show Routed minus Shared in percentage points; bars are 95\% paired-bootstrap confidence intervals over 200 matched examples.}
\label{fig:paired-routing-effects}
\end{figure}

Table~\ref{tab:paired-routing-effects} reports Shared and Routed scores together with 95\% confidence intervals (CIs), differences in percentage points (pp), and per-example win/tie/loss (W/T/L) counts over the same 200 examples.  Figure~\ref{fig:paired-routing-effects} visualizes the corresponding differences.  The intervals show a small negative shift on MuSiQue and a small positive shift on 2WikiMultiHopQA; neither is statistically reliable across all metrics.  HotpotQA is analyzed separately in Appendix~\appjump{app.hotpot-fullwiki-stress}{D.6}.

\subsection{Role Induction and Evaluation Diagnostics}
\appanchor{app.role-routing-diagnostics}

Figure~\ref{fig:app-role-evaluation-diagnostics} jointly examines role-count sensitivity, cross-seed stability, and paired trajectory changes.  Panel~(a) shows that $K=3$ remains competitive in silhouette, within-run stability, and support entropy, whereas $K=4$ and $K=5$ achieve slightly stronger clustering geometry.  We therefore treat three roles as a compact, team-aligned operating point, not as the uniquely optimal partition.  Panel~(b) evaluates three $K=3$ induction runs on the same 323 trajectory prefixes: adjusted Rand index (ARI) ranges from 0.61 to 0.85, normalized mutual information (NMI) from 0.62 to 0.82, and Hungarian-aligned profile cosine from 0.80 to 0.97.  Panel~(c) provides a qualitative view of the largest positive and negative paired MuSiQue changes.  These selected cases do not estimate the population-level routing effect; instead, they show that routing can alter both retrieval decisions and evidence-to-answer conversion.  Taken together, the diagnostics support moderately stable role induction while showing that downstream routing effects remain example dependent.

\subsection{Role-Agent-Turn Disentanglement}
\appanchor{app.role-agent-turn-disentanglement}

We test whether an induced role captures a transferable behavior or merely renames a fixed agent index or speaking position.  On the MuSiQue validation set, we independently permute three speaker orders and three agent-role assignments for both ExRole-Shared and ExRole-Routed.  The resulting $3\times3\times2$ design contains 18 conditions and 3,600 episodes; within each variant, every role is enacted by every agent in early, middle, and late phases.  This is an inference-time intervention on fixed checkpoints, not an additional training run.

Table~\ref{tab:role-agent-turn-disentanglement} reports two complementary tests.  First, adding role identity to an action model that already observes agent identity and phase reduces held-out log loss by 19.14\% for Shared and 18.68\% for Routed, while increasing macro-F1 by approximately 0.236.  Second, the action distribution of the same role changes little across agents, whereas the mean distance between different roles is 48--60 times larger.  Role identity therefore explains behavior beyond agent index and coarse turn phase.  The schedule-level results also preserve the main Routed-versus-Shared finding: routing does not improve the average score under these interventions.

\begin{table}[H]
\centering
\scriptsize
\setlength{\tabcolsep}{2.3pt}
\renewcommand{\arraystretch}{1.08}
\begin{tabular*}{\columnwidth}{@{\extracolsep{\fill}}lccc@{}}
\toprule
\textbf{Measure} & \textbf{Shared} & \textbf{Routed} & \textbf{Effect} \\
\midrule
Agent + phase log loss & 0.4439 & 0.4408 & control \\
+ Role log loss & 0.3589 & 0.3585 & $-19.14\%/-18.68\%$ \\
+ Role macro-F1 & 0.6809 & 0.6808 & $\approx+0.236$ \\
Same-role cross-agent JS & 0.00446 & 0.00550 & low variation \\
Different-role JS & 0.2680 & 0.2652 & $60.1\times/48.2\times$ \\
\rowcolor{bdexroleGray}
Mean EM over nine schedules & 29.22 & 28.33 & $-0.89$ pp \\
\rowcolor{bdexroleGray}
Mean F1 over nine schedules & 40.99 & 39.97 & $-1.02$ pp \\
\bottomrule
\end{tabular*}
\caption{Role-Agent-Turn disentanglement on MuSiQue.  Action prediction uses 8,429 Shared and 8,423 Routed turns; schedule-level scores average nine orthogonal conditions per checkpoint.  JS denotes Jensen--Shannon distance.}
\label{tab:role-agent-turn-disentanglement}
\end{table}

\begin{figure}[t]
\centering
\includegraphics[width=\linewidth]{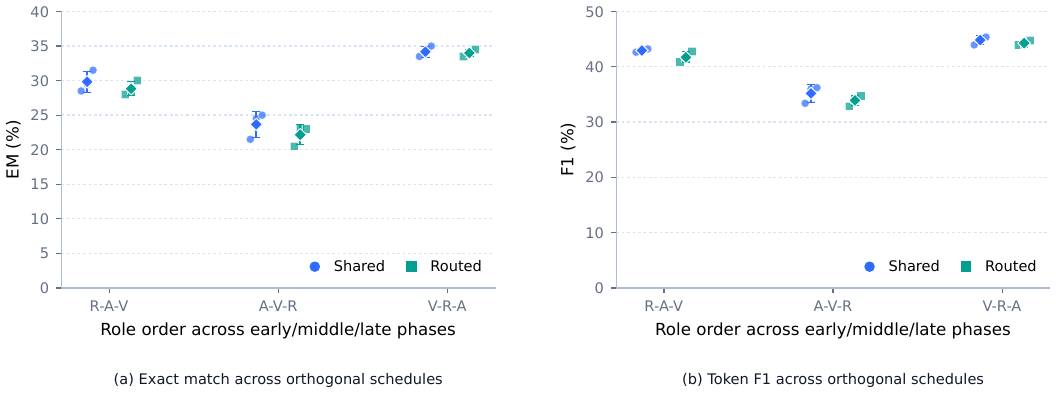}
\caption{Role-stage effects under orthogonal MuSiQue schedules.  Small points are the three speaker-order and agent-role-map conditions associated with each role-stage pattern; diamonds and bars show the mean and one sample standard deviation.  R, A, and V denote Researcher, Analyst, and Verifier.}
\label{fig:role-agent-turn-phase-effect}
\end{figure}

Figure~\ref{fig:role-agent-turn-phase-effect} isolates the position of each role in the early, middle, and late phases while retaining all three orthogonal scheduling points.  Moving the Analyst--Verifier--Researcher order to these phases lowers both EM and F1, whereas Verifier--Researcher--Analyst is strongest.  Thus a role is not reducible to a fixed turn label, but its utility still depends on when it acts in the collaboration sequence.

\begin{figure}[t]
\centering
\includegraphics[width=\linewidth]{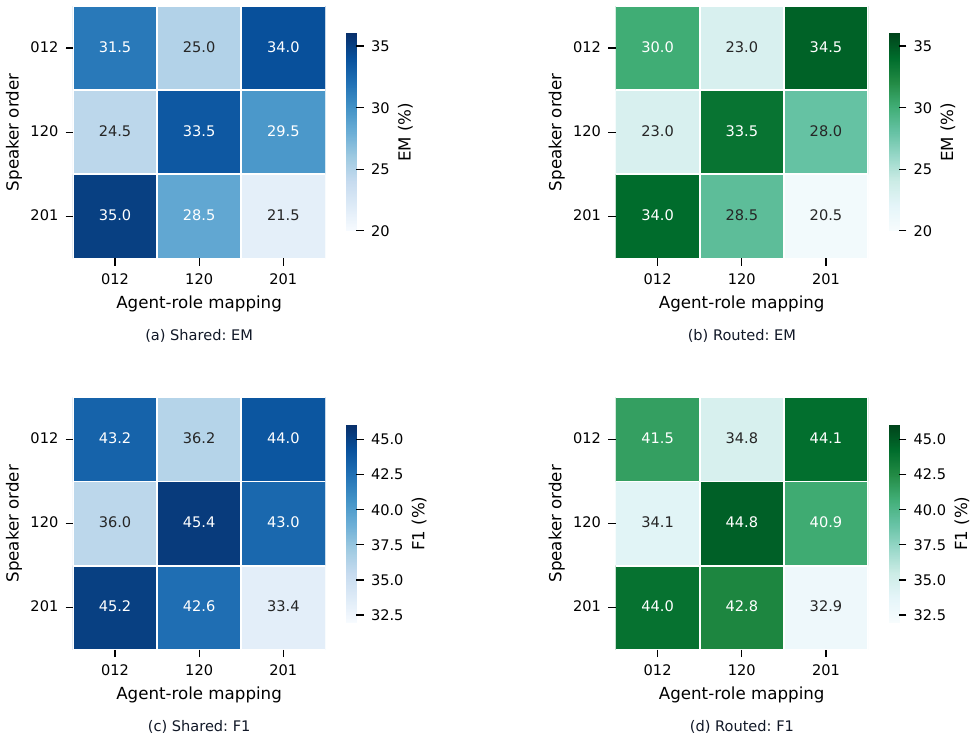}
\caption{Complete Role-Agent-Turn intervention matrix on MuSiQue.  Rows vary speaker order and columns vary agent-role assignment; cells report EM or F1 for the same 200 questions.  Shared and Routed use identical schedules and evaluation protocol.}
\label{fig:role-agent-turn-schedule-matrix}
\end{figure}

Figure~\ref{fig:role-agent-turn-schedule-matrix} gives the complete $3\times3$ schedule matrix rather than only phase-aggregated values.  High- and low-performing cells occur under multiple speaker orders and agent-role mappings, which rules out a single privileged agent assignment.  At the same time, the structured variation across cells confirms that coordination order remains part of the task rather than a nuisance variable that can be ignored.

\Needspace{14\baselineskip}
\subsection{Complementarity of Role Induction and Turn-Aligned Credit}
\appanchor{app.role-credit-complementarity}

We isolate the two learning components with an independently trained $2\times2$ design under the shared-capacity setting.  Among the four matched runs, jointly enabling trajectory-induced roles and turn-aligned credit achieves the highest observed result, reaching 30.0 EM and 41.5 F1.  Relative to this joint configuration, removing role induction reduces EM and F1 by 3.5 and 2.3 points, respectively, while removing turn-aligned credit decreases F1 by 2.0 points.  This pattern is consistent with complementary functions: induced role structure organizes collaborative behavior, while turn-aligned credit connects that structure to policy updates.

\begin{table}[H]
\centering
\scriptsize
\setlength{\tabcolsep}{2.6pt}
\renewcommand{\arraystretch}{1.10}
\begin{tabular*}{\columnwidth}{@{\extracolsep{\fill}}lcccc@{}}
\toprule
\textbf{Configuration} & \textbf{Role} & \textbf{Credit} & \textbf{EM \metricup} & \textbf{F1 \metricup} \\
\midrule
\rowcolor{bdexroleGray}
\textbf{Full role-credit} & \textcolor{bdgreen}{\ding{51}} & \textcolor{bdgreen}{\ding{51}} & \bestplain{30.0} & \bestplain{41.5} \\
w/o role induction & \textcolor{bdneutral}{\ding{55}} & \textcolor{bdgreen}{\ding{51}} & \mneg{26.5}{-3.5} & \mneg{39.2}{-2.3} \\
w/o turn-aligned credit & \textcolor{bdgreen}{\ding{51}} & \textcolor{bdneutral}{\ding{55}} & \mneg{29.5}{-0.5} & \mneg{39.5}{-2.0} \\
w/o both & \textcolor{bdneutral}{\ding{55}} & \textcolor{bdneutral}{\ding{55}} & \mneg{29.0}{-1.0} & \mneg{40.3}{-1.2} \\
\bottomrule
\end{tabular*}
\caption{Strict $2\times2$ component ablation on MuSiQue validation-200.  The four variants are independently trained with seed 42 and share the same backbone, 128-update horizon, and evaluation protocol.  Parenthesized values are changes relative to the complete role-credit configuration.}
\label{tab:role-credit-2x2}
\end{table}

\subsection{Inference Cost and Answer Diagnostics}
\appanchor{app.inference-diagnostics}

Table~\ref{tab:inference-answer-diagnostics} separates computational budget from answer quality on the same 200 MuSiQue examples.  Shared and Routed have nearly identical turn, search, retrieval, and repetition statistics.  Target-bearing evidence is available substantially more often than a strictly correct answer is produced, identifying final answer formation as a major remaining bottleneck.

\subsection{Full-Wikipedia HotpotQA Stress Test}
\appanchor{app.hotpot-fullwiki-stress}

HotpotQA differs from the two primary benchmarks by retrieving from a lexical full-Wikipedia index rather than a bounded supporting-document collection.  We therefore report it as a retrieval stress test rather than use it as primary evidence for the role-learning claim.  The team interface still uses three agents, six turns, top-3 retrieval, and the same EM/F1/Succ evaluation implementation.

\begin{table}[H]
\centering
\scriptsize
\setlength{\tabcolsep}{3.2pt}
\renewcommand{\arraystretch}{1.08}
\begin{tabular*}{\columnwidth}{@{\extracolsep{\fill}}lccc@{}}
\toprule
\textbf{Method} & \textbf{EM \metricup} & \textbf{F1 \metricup} & \textbf{Succ. \metricup} \\
\midrule
Single-agent search & 27.0 & 36.0 & 27.0 \\
\midrule
No-role MAS & 33.5 & \bestplain{44.4} & 33.0 \\
Manual-role MAS & 33.0 & 42.9 & 32.5 \\
Random role prompt & \bestplain{34.5} & 43.9 & \bestplain{34.0} \\
Shuffled induced role & 33.5 & 42.4 & 33.5 \\
\midrule
\rowcolor{bdexroleGray}
ExRole-Shared & 30.0 & 38.3 & 30.0 \\
\rowcolor{bdexroleGray}
ExRole-Routed & 30.5 & 37.8 & 30.5 \\
\bottomrule
\end{tabular*}
\caption{HotpotQA results under the full-Wikipedia FTS5 retrieval stress test.  Succ. is strict success, computed independently from EM.}
\label{tab:hotpot-fullwiki-stress}
\end{table}

Neither ExRole variant surpasses the strongest role-free or prompt-role controls in this setting.  Routed changes Shared by $+0.5$ EM with a 95\% CI of $[0.0,+1.5]$ and by $-0.5$ F1 with a CI of $[-1.4,+0.3]$, indicating negligible routing sensitivity.  These results establish a boundary of the current method: role specialization alone does not overcome the retrieval and answer-selection constraints of this full-Wikipedia configuration.  They do not, by themselves, isolate retrieval quality as the causal source of the gap.

\begin{figure}[H]
\centering
\begin{minipage}[t]{0.485\columnwidth}
  \centering\vspace{0pt}
  \includegraphics[width=\linewidth]{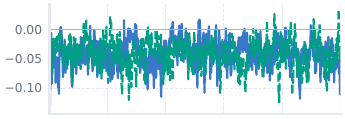}\par
  \vspace{-1.0mm}
  {\scriptsize\textbf{(a)} Shared/Routed policy loss.\par}
\end{minipage}\hfill
\begin{minipage}[t]{0.485\columnwidth}
  \centering\vspace{0pt}
  \includegraphics[width=\linewidth]{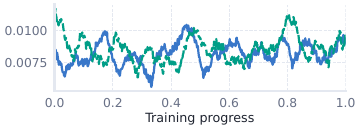}\par
  \vspace{-1.0mm}
  {\scriptsize\textbf{(b)} Shared/Routed turn credit.\par}
\end{minipage}
\vspace{1.2mm}

\includegraphics[width=0.72\columnwidth]{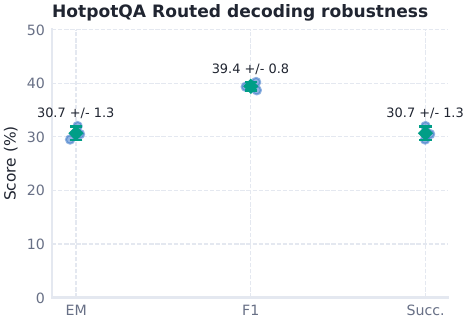}\par
\vspace{-1.0mm}
{\scriptsize\textbf{(c)} Routed decoding robustness over three seeds.\par}
\caption{Optimization and decoding diagnostics for the HotpotQA full-Wikipedia stress test.  Panels (a) and (b) use the matched 128-update Shared/Routed runs; panel (c) reports three temperature-1 evaluations of a fixed Routed checkpoint on the same 200 questions.}
\label{fig:hotpot-fullwiki-stress}
\end{figure}

\begin{table}[H]
\centering
\scriptsize
\setlength{\tabcolsep}{3.2pt}
\renewcommand{\arraystretch}{1.08}
\begin{tabular*}{\columnwidth}{@{\extracolsep{\fill}}lccc@{}}
\toprule
\textbf{Method} & \textbf{Base} & \textbf{EM \metricup} & \textbf{F1 \metricup} \\
\midrule
\multicolumn{4}{@{}l}{\textit{Retrieval baselines}} \\
Direct & \basesft & 16.5 & 24.6 \\
Naive RAG & \basesft & 20.5 & 26.3 \\
\midrule
\multicolumn{4}{@{}l}{\textit{Search-RL baselines}} \\
Search-R1 & \baseqwen & 27.5 & 35.2 \\
R1-Searcher & \baseragrl & 27.5 & 35.4 \\
\midrule
\multicolumn{4}{@{}l}{\textit{MAS-RL baselines}} \\
MAGRPO & \basesft & 23.0 & 29.2 \\
Dr. MAS & \basesft & 27.0 & 34.4 \\
MATPO & \basesft & 33.5 & 43.7 \\
\midrule
\rowcolor{bdexroleGray}
ExRole-Shared & \basesft & 30.0 & 38.3 \\
\rowcolor{bdexroleGray}
ExRole-Routed & \basesft & 30.5 & 37.8 \\
\bottomrule
\end{tabular*}
\caption{HotpotQA external references under each method's stated protocol.  Retrieval, training, and evaluation details differ across rows, so the values provide context rather than a controlled ranking.}
\label{tab:hotpot-external-reference}
\end{table}

\subsection{External Reference Training Curves}
\appanchor{app.external-baseline-curves}

We adapt MAGRPO and Dr. MAS to MuSiQue, HotpotQA, and 2WikiMultiHopQA with the same Qwen2.5-7B SFT policy backbone.  Figure~\ref{fig:app-external-training-curves} combines their learning dynamics: MAGRPO uses a sequential three-agent search interface and grouped updates over 512 training instances, while Dr. MAS uses shared-model grouped rollouts over 128 updates with an 8192-token context budget.  Panel~(a) reports MAGRPO reward, policy loss, and final 200-example EM/F1; panel~(b) reports Dr. MAS reward, valid-action ratio, and native periodic validation.  The horizontal axis is normalized within each run, so the curves verify non-degenerate optimization but do not equate method-specific objectives or logging schedules.

\makeatletter
\setlength{\@dblfptop}{0pt}
\setlength{\@dblfpbot}{0pt plus 1fil}
\makeatother

\begin{table*}[!t]
\centering
\scriptsize
\setlength{\tabcolsep}{2.5pt}
\renewcommand{\arraystretch}{1.14}
\begin{tabular*}{\textwidth}{@{\extracolsep{\fill}}llcccc@{}}
\toprule
\textbf{Diagnostic group} & \textbf{Quantity} & \textbf{Unit} & \textbf{Shared} & \textbf{Routed} & \textbf{$\Delta$} \\
\midrule
& Team turns & count & $4.48\pm1.50;\ 3.0[3.0,6.0]$ & $4.49\pm1.50;\ 4.0[3.0,6.0]$ & \textcolor{bdneutral}{$+0.02$} \\
& Search calls & count & $3.48\pm1.50;\ 2.0[2.0,5.0]$ & $3.49\pm1.50;\ 3.0[2.0,5.0]$ & \textcolor{bdneutral}{$+0.02$} \\
& Retrieved text & kchars & $5.35\pm3.42;\ 4.50[2.52,7.30]$ & $5.35\pm3.34;\ 4.56[2.53,7.41]$ & \textcolor{bdneutral}{$-0.01$} \\
\multirow{-4}{*}{\textbf{Resource use}} & Repeated queries & count & $0.33\pm0.80;\ 0.0[0.0,0.0]$ & $0.36\pm0.83;\ 0.0[0.0,0.0]$ & \textcolor{bdneutral}{$+0.03$} \\
\midrule
& Search issued & episodes (\%) & 100.0 & 100.0 & \textcolor{bdneutral}{$0.0$} \\
& Target retrieved & episodes (\%) & 73.5 & 73.0 & \textcolor{bdred}{$-0.5$} \\
& Gold span visible & episodes (\%) & 73.5 & 73.5 & \textcolor{bdneutral}{$0.0$} \\
& Grounded answer & episodes (\%) & 87.0 & 87.0 & \textcolor{bdneutral}{$0.0$} \\
\multirow{-5}{*}{\textbf{Answer stages}} & Strict success & episodes (\%) & 31.5 & 30.0 & \textcolor{bdred}{$-1.5$} \\
\bottomrule
\end{tabular*}
\caption{Matched MuSiQue inference and answer diagnostics over the same 200 examples. Resource-use cells report mean$\pm$standard deviation (SD) followed by median [interquartile range (IQR)]; stage rows report percentages. $\Delta$ is ExRole-Routed minus ExRole-Shared. Stage conditions are evaluated independently and do not form a monotone funnel.}
\label{tab:inference-answer-diagnostics}
\vspace{3.0mm}

\begin{minipage}[t]{\textwidth}
  \centering\vspace{0pt}
  \includegraphics[width=\linewidth]{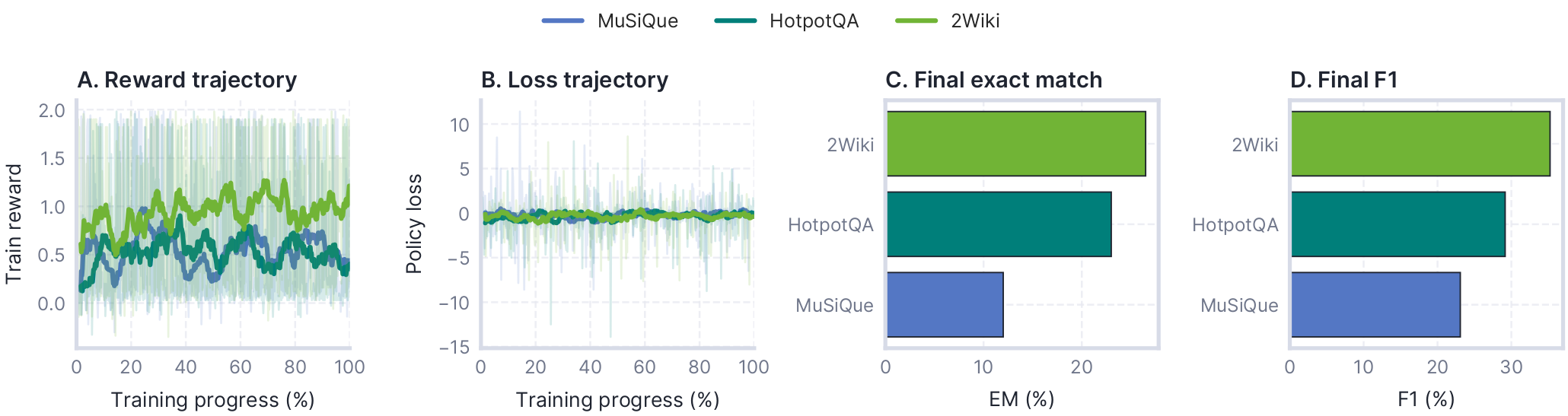}\par
  \vspace{-1.0mm}
  {\small\textbf{(a)} MAGRPO training dynamics and final evaluation.\par}
\end{minipage}
\vspace{2.0mm}

\begin{minipage}[t]{\textwidth}
  \centering\vspace{0pt}
  \includegraphics[width=\linewidth]{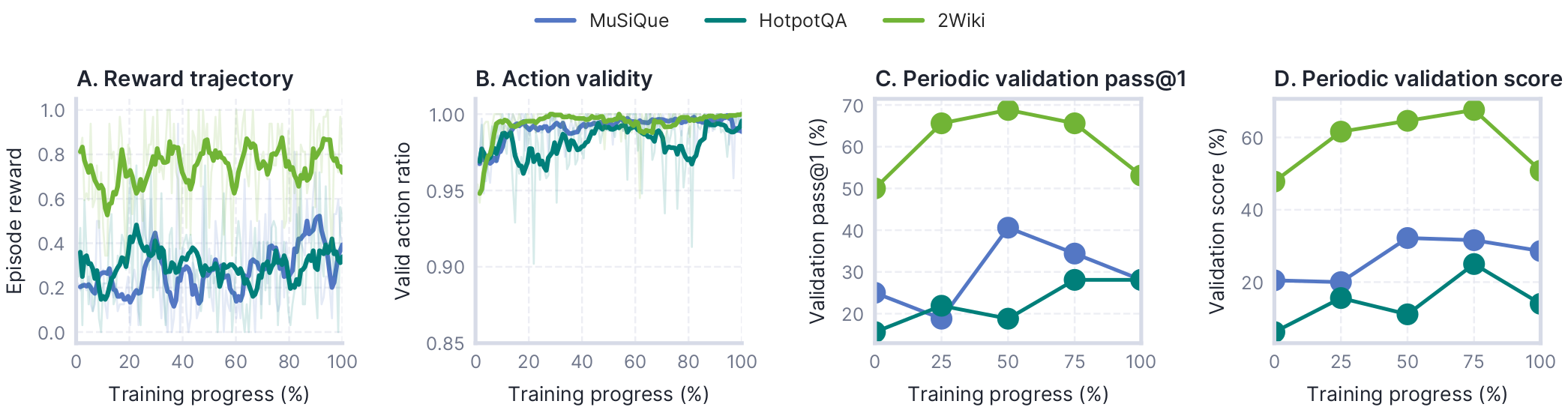}\par
  \vspace{-1.0mm}
  {\small\textbf{(b)} Dr. MAS training dynamics and periodic validation.\par}
\end{minipage}
\captionof{figure}{External MAS-RL baseline reproductions on MuSiQue, HotpotQA, and 2WikiMultiHopQA. Both panels use normalized training progress on the horizontal axis. MAGRPO reports reward, policy loss, and final 200-example EM/F1, while Dr. MAS reports reward, valid-action ratio, and its native periodic validation measurements.}
\label{fig:app-external-training-curves}
\end{table*}

\begin{table*}[!t]
\centering
\small
\setlength{\tabcolsep}{5pt}
\renewcommand{\arraystretch}{1.12}
\begin{tabular}{@{}L{0.12\textwidth}C{0.055\textwidth}C{0.085\textwidth}C{0.085\textwidth}C{0.085\textwidth}C{0.16\textwidth}C{0.09\textwidth}@{}}
\toprule
\textbf{Benchmark} & \textbf{Metric} & \multicolumn{2}{c}{\textbf{Score}} & \multicolumn{3}{c}{\textbf{Paired Routed $-$ Shared effect}} \\
\cmidrule(lr){3-4}\cmidrule(lr){5-7}
& & \textbf{Shared} & \textbf{Routed} & \textbf{$\Delta$ (pp)} & \textbf{95\% CI} & \textbf{W/T/L} \\
\midrule
& EM & 31.5 & 30.0 & \cellcolor{bdsoftred}\textcolor{bdred}{\textbf{$-1.5$}} & $[-4.0,\,+0.5]$ & 1/195/4 \\
& F1 & 43.2 & 41.5 & \cellcolor{bdsoftred}\textcolor{bdred}{\textbf{$-1.8$}} & $[-4.2,\,+0.3]$ & 2/193/5 \\
\multirow{-3}{*}{\textbf{MuSiQue}} & Succ. & 31.5 & 30.0 & \cellcolor{bdsoftred}\textcolor{bdred}{\textbf{$-1.5$}} & $[-4.0,\,+0.5]$ & 1/195/4 \\
\midrule
& EM & 49.0 & 50.0 & \cellcolor{bdtealWin}\textcolor{bdgreen}{\textbf{$+1.0$}} & $[-1.5,\,+3.5]$ & 4/194/2 \\
& F1 & 59.1 & 59.7 & \cellcolor{bdtealWin}\textcolor{bdgreen}{\textbf{$+0.7$}} & $[-0.8,\,+2.3]$ & 5/191/4 \\
\multirow{-3}{*}{\textbf{2Wiki}} & Succ. & 49.0 & 50.0 & \cellcolor{bdtealWin}\textcolor{bdgreen}{\textbf{$+1.0$}} & $[-1.5,\,+3.5]$ & 4/194/2 \\
\bottomrule
\end{tabular}
\caption{Paired Shared-versus-Routed comparison over 200 examples for each primary benchmark. $\Delta$ is measured in percentage points; intervals are 95\% paired-bootstrap confidence intervals with 20,000 resamples. W/T/L counts per-example Routed wins, ties, and losses.}
\label{tab:paired-routing-effects}
\vspace{3.0mm}

\begin{minipage}[t]{0.32\textwidth}
  \centering\vspace{0pt}
  \includegraphics[width=\linewidth]{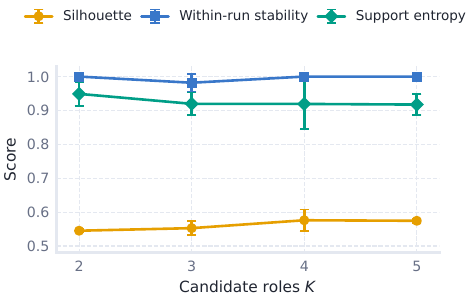}\par
  \vspace{-1.0mm}
  {\small\textbf{(a)} Role-count sensitivity over three discovery seeds.\par}
\end{minipage}\hfill
\begin{minipage}[t]{0.32\textwidth}
  \centering\vspace{0pt}
  \includegraphics[width=0.84\linewidth]{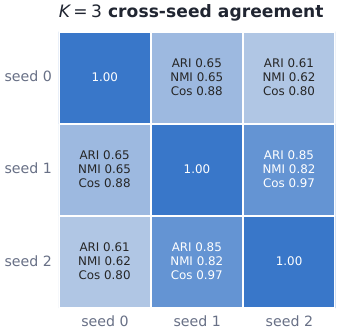}\par
  \vspace{-1.0mm}
  {\small\textbf{(b)} Cross-seed role agreement for $K=3$.\par}
\end{minipage}
\hfill
\begin{minipage}[t]{0.32\textwidth}
  \centering\vspace{0pt}
  \includegraphics[width=\linewidth]{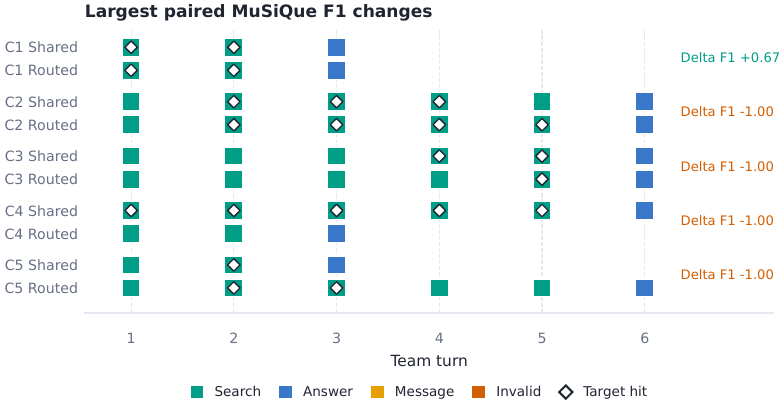}\par
  \vspace{-1.0mm}
  {\small\textbf{(c)} Extreme paired MuSiQue trajectory changes.\par}
\end{minipage}
\captionof{figure}{Role-induction and evaluation diagnostics. Panel (a) evaluates sensitivity to the number of induced roles; panel (b) measures cross-seed agreement of the induced partition and role profiles; and panel (c) traces paired examples with the largest Routed-versus-Shared F1 changes. Error bars denote one sample standard deviation where applicable.}
\label{fig:app-role-evaluation-diagnostics}
\end{table*}

\begin{table*}[!t]
\section{Notation}
\subsection{Symbol Table}
\appanchor{app.symbol-table}
\centering
\footnotesize
\setlength{\tabcolsep}{3pt}
\renewcommand{\arraystretch}{1.10}
\begin{tabular*}{\textwidth}{@{\extracolsep{\fill}}p{0.14\textwidth}p{0.32\textwidth}p{0.14\textwidth}p{0.32\textwidth}@{}}
\toprule
\textbf{Symbol} & \textbf{Meaning} & \textbf{Symbol} & \textbf{Meaning} \\
\cmidrule(r){1-2}\cmidrule(l){3-4}
$x,y,\tau$ & Task input, gold answer, and team trajectory & $S,L,d$ & Slot budget, adapted linear modules, and LoRA rank \\
$N$ & Number of agents in the team & $s$ & LoRA rank-slot index, $1\leq s\leq d$ \\
$T,T_\tau$ & Maximum and realized numbers of team turns & $\omega_{q,\ell s}$ & Score for module $\ell$ and rank slot $s$ in segment $q$ \\
$J,K$ & Numbers of induction records and source clusters & $\varpi_{q,\ell s}$ & Normalized routing mass over LoRA rank slots \\
$G,Q$ & GRPO group size and number of role-turn segments & $\tilde{b}_{q,\ell s}$ & Soft slot budget before hard top-$S$ selection \\
$i_t$ & Active speaker at turn $t$ & $M_{q,\ell s}$ & Hard top-$S$ slot mask \\
$o_t$ & Text observation given to the active agent & $\gamma_{q,\ell s}$ & Effective LoRA multiplier after sparse-delta routing \\
$a_t,m_t$ & Structured action and optional teammate message & $c_q,\widehat c_q$ & Detached turn-credit target and router prediction \\
$e_t$ & Retrieved evidence after a search action & $\mathcal B_R$ & Role-turn segments in the current normalization batch \\
$z_i,z_q,z(u)$ & Runtime role of an agent, segment, or token & $\theta,\theta_e,\theta_b$ & Policy, role-encoder, and router parameters \\
$\xi_t,\phi_t^{\mathrm{obs}}$ & Learned role embedding and prefix-local behavior feature vector & $\tau_b,\lambda_{\mathrm{disc}}$ & Routing temperature and turn-credit discount \\
$p_k$ & Induced source prototype for cluster $k$ & $\mathrm{Adv}^{(g)},\rho_{t,j}^{(g)}$ & Group advantage and token probability ratio \\
$\eta_k$ & Action, stage, evidence, and return statistics for cluster $k$ & $U_t^{\mathrm{turn}},\alpha_c$ & Discounted turn return and policy-credit mixing weight \\
$\psi_z,\chi_z$ & Readable role instruction and token-aligned role marker & $R(\tau,y)$ & Episode reward against gold answer $y$ \\
$q,\operatorname{seg}(u)$ & Role-turn segment and token-to-segment map & $n_{\mathrm{em}},n_{\mathrm{succ}}$ & Benchmark and strict environment normalizers \\
$\Gamma_t,\Gamma_q,\Gamma(u)$ & Adapter-routing state at turn, segment, or token level & $N_{\mathrm{eval}},\mathrm{Succ}$ & Evaluation-set size and strict success rate \\
$\mathbf{v}_{z_q},\zeta_q$ & Executable-role feature and pooled semantic prefix & & \\
\bottomrule
\end{tabular*}
\caption{Symbols used throughout the method, training objective, and evaluation protocol.}
\label{tab:symbol-table}
\end{table*}

\end{document}